%% file: main.tex
\documentclass[twoside]{article}
\usepackage{PRIMEarxiv}
\pdfoutput=1
\usepackage[frozencache,cachedir=.]{minted}  
\input{macros}

\usepackage[utf8]{inputenc} 
\usepackage[T1]{fontenc}    
\usepackage{hyperref}       
\usepackage{url}            
\usepackage{booktabs}       
\usepackage{amsfonts}       
\usepackage{nicefrac}       
\usepackage{microtype}      
\usepackage{lipsum}
\usepackage{fancyhdr}       
\usepackage{graphicx}       
\graphicspath{{media/}}     

\usepackage[sort&compress,square,comma,authoryear]{natbib}

\title{Semi-Equivariant Continuous Normalizing Flows for Target-Aware Molecule Generation}

\author{
  Eyal Rozenberg, Daniel Freedman \\
  Verily Research\\
  Haifa, Israel \\
  \texttt{\{eyalrozenberg,danielfreedman\}@verily.com}
}

\begin{document}
\maketitle

\begin{abstract}
\input{text/abstract}
\end{abstract}


\section{Introduction}
\input{text/introduction}

\section{Related Work}
\input{text/related_work}

\section{Invariant Conditional Model of Ligands Given a Receptor}
\input{text/method}

\section{Experiments}
\label{sec:experiments}
\input{text/experiments}

\section{Conclusions}
\input{text/conclusions}

\bibliographystyle{plainnat}
\bibliography{references}  

\appendix
\section{Appendix}
\input{text/appendix}

\end{document}

%% file: macros.tex
\usepackage{amsmath,amsthm,amssymb,amsfonts}
\usepackage{booktabs,array,tabularx}

\usepackage{physics}
\usepackage{multirow}
\usepackage{esvect}
\usepackage{ulem}
\usepackage{cancel}

\usepackage{dsfont}
\usepackage{subcaption}
\usepackage{bm}
\usepackage{svg}

\usepackage[skins]{tcolorbox}
\usepackage{tikz}
\usetikzlibrary{
    shapes, 
    arrows.meta, 
    positioning}

\usepackage{pgfplots}
\pgfplotsset{width=10cm,compat=1.9}

\usepackage{algorithm}
\usepackage{paralist}
\usepackage{minted}

\newtheorem*{theorem}{Theorem}
\newtheorem{lemma}{Lemma}

\newcommand{\fig}[2]{\includegraphics[width=#2\textwidth]{imgs/#1}}
\newcommand{\parbasic}[1]{\textbf{#1} \hspace{0.5mm}}

\newcommand{\bh}[1]{\mathbf{\hat{#1}}}
\newcommand{\bt}[1]{\mathbf{\tilde{#1}}}

%% file: text/abstract.tex
We propose an algorithm for learning a conditional generative model of a molecule given a target.  Specifically, given a receptor molecule that one wishes to bind to, the conditional model generates candidate ligand molecules that may bind to it.  The distribution should be invariant to rigid body transformations that act \textit{jointly} on the ligand and the receptor; it should also be invariant to permutations of either the ligand or receptor atoms.  Our learning algorithm is based on a continuous normalizing flow.  We establish semi-equivariance conditions on the flow which guarantee the aforementioned invariance conditions on the conditional distribution.  We propose a graph neural network architecture which implements this flow, and which is designed to learn effectively despite the vast differences in size between the ligand and receptor.  We evaluate our method on the CrossDocked2020 dataset, attaining a significant improvement in binding affinity over competing methods.

%% file: text/introduction.tex
The design of new molecules is an important topic, with applications in medicine, biochemistry, and materials science.  Recently there have been quite a number of promising directions for applying machine learning techniques to this problem; most relevant to the current work are those which produce a generative model of molecules.  To date, the majority of approaches have focused on the \textit{unconditional} setting, in which the goal is simply to produce molecules without regard for a more specific purpose.  Such techniques can, for example, effectively produce ``drug-like'' molecules.  However, given two different targets that one wishes to bind to, the models will produce precisely the same distribution of candidates in both cases.

In this paper, we focus on learning generative models for molecules in the \textit{conditional} setting.  Specifically, we assume that we are given a target receptor molecule; the aim is then to generate ligand molecules that may successfully bind to this receptor.  This kind of conditional generative model is very useful in the context of drug design, in which one often has a target (receptor) in mind, and the goal is then to find drugs (ligands) which will bind to the target.

It is important that our generative model be probabilistic, so that we can generate multiple potential candidates; diversity is useful in this context, as not all candidates will be equally suitable experimentally, due to considerations such as toxicity.  We choose to describe both the receptor and ligand as 3D graphs, so our goal is to learn a conditional probabilistic model of one 3D graph given another.

We implement the central part of this conditional probabilistic model as a continuous normalizing flow.  The work of \citet{satorras2021flows} pioneered this approach in the unconditional setting.  In the conditional setting, a number of important changes must be made, and the conditioning variable (the receptor) must enter the flow in a very particular way.  Specifically, we know that the probabilistic model must capture natural invariances, to rigid motions and permutations of the atoms.  Note that the invariance to rigid motions is a kind of ``conditional invariance'', which is expressed \textit{jointly} in terms of the ligand and the receptor.  Due to the fact that the receptor is a conditioning variable, this leads to a novel form of \textit{semi-equivariance} of the flow, which we prove.  An additional issue which arises is the vastly different sizes of the two molecules: the receptor is typically 1-2 orders of magnitude larger than the ligand.  We introduce an architecture for implementing the flow which takes this into account, and enables effective learning despite the size disparity.  We train our method on the CrossDocked2020 dataset \citep{francoeur2020three} and attain high quality performance, with a significant improvement over competing methods in the key $\Delta$Binding metric.

Our principal contributions are as follows:
\begin{itemize}
    \item The derivation of conditions on a continuous normalizing flow which allows for joint invariance of the ligand and receptor to both rigid motions and permutations.
    \item The design of an architecture which implements the conditional distribution of ligand given receptor, enabling effective learning despite the size disparity between the molecules.
    \item A demonstration of the effectiveness of the method on the CrossDocked2020 dataset, attaining a significant improvement versus competitors in terms of $\Delta$Binding.
\end{itemize}

%% file: text/related_work.tex
\parbasic{Continuous normalizing flows} Normalizing flows transform an initial density distribution into a desired density distribution by applying a sequence of invertible transformations.  They are easy to sample from and can be trained by maximum likelihood using the change of variables formula \citep{rippel2013high, dinh2014nice, rezende2015variational, dinh2016density, kingma2018glow}. \citet{chen2018continuous} introduced a continuous flow based on ODEs, which was then extended in \citep{grathwohl2018ffjord}. Further works aimed to improve the accuracy of the model \citep{zhuang2021mali, zhang2022pnode}; improve stability and deal with ODE stiffness issues through regularization \citep{finlay2020train, kelly2020learning, ghosh2020steer}; and deal with discrete data \citep{ho2019flow++,hoogeboom2021argmax}.

\parbasic{GNNs for drug design and discovery} Graph Neural Networks (GNNs) are well-suited to problems in molecular modelling, with atoms and bonds represented as vertices and edges, respectively.  GNN methods were shown to be useful for detecting receptor's binding site (pocket) and predicting the receptor pocket-ligand or protein-protein interactions \citep{gainza2020deciphering, sverrisson2021fast, sverrisson2022physics}; as well as the drug binding structure or the rigid body protein-protein docking structure \citep{ganea2021independent, stark2022equibind}. Several generative methods to produce molecules were previously demonstrated, for the unconditional setting \citep{gebauer2019symmetry, satorras2021flows, hoogeboom2022equivariant, trippe2022diffusion}. \citet{satorras2021flows} constructed an E(n) equivariant normalizing flows, by incorporating equivariant graph neural networks \citep{satorras2021egnn} into an ODE framework to obtain an invertible equivariant function. The model to jointly generates molecular features and 3D positions.

\parbasic{3D molecular design conditioned on a receptor binding site}
This line of research is newer than the general (unconditioned) GNN-based methods described above. LiGAN \citep{ragoza2022generating} is a pioneering work in this area, which uses a condition VAE on an image-like discretized 3D atomic density grid; a post-sampling step converts this grid structure into molecules using an atom fitting algorithm.  A more recent class of works \citep{luo20213d, liu2022generating} use an auto-regressive approach.
\citet{luo20213d} derive a model which captures the probability that a point 3D space is occupied by an atom of a particular chemical element. \citet{liu2022generating} propose the GraphBP framework, which eliminates the need to discretize space; to place a new atom, they generate its atom type and relative location while preserving the equivariance property.  Other works include fragment-based ligand generation, in which new molecular fragments are sequentially attached to the growing molecule \citep{powers2022fragment}; abstraction of the geometric interaction features of the receptor–ligand complex to a latent space, for generative models such as Bayesian sampling \citep{wang2022relation} and RNNs \citep{zhang2022novo}; and use of experimental electron densities as training data for the conditional generative model \citep{wang2022pocket}.

\parbasic{Data} Datasets with a large number of receptor-ligand complexes are critical to our endeavour.  Many models have relied on the high quality PDBbind dataset which curates the Protein Data Bank (PDB) \citep{liu2017forging}; however, for the training of generative models, this dataset is relatively small.  CrossDocked2020 \citep{francoeur2020three} is the first large-scale standardized dataset for training ML models with ligand poses cross-docked against non-cognate receptor structure, greatly expanding the number of poses available for training. The dataset is organized by clustering of similar binding pockets across the PDB; each cluster contains ligands cross-docked against all receptors in the pocket. Each receptor-ligand structure also contains information indicating the nature of the docked pair, such as root mean squared deviation (RMSD) to the reference crystal pose and Vina cross-docking score \citep{trott2010autodock} as implemented in Smina \citep{koes2013lessons}. The dataset contains 22.5 million poses of ligands docked into multiple similar binding pockets across the PDB.

%% file: text/method.tex
\parbasic{Objective}
Our overall goal can be stated as follows: wish to learn a conditional distribution over ligand molecules given a particular receptor molecule.  As both ligand and receptor are described by 3D graphs, this will be a distribution of the form $p(\mathcal{G} | \hat{\mathcal{G}})$.  To be physically plausible, this distribution must be invariant to certain groups of transformations: rigid motions applied jointly the entire ligand-receptor complex, as well as permutations applied to each of the ligand and receptor separately.

\parbasic{Organization}
This section proposes a method for learning such a conditional distribution $p(\mathcal{G} | \hat{\mathcal{G}})$.  Section \ref{sec:notation} introduces notation.  Section \ref{sec:decomposition} decomposes the distribution into 4 subdistributions using a Markov decomposition, and shows how the invariance properties apply to each subdistribution.  Section \ref{sec:intermezzo} is an intermezzo, describing two types of Equivariant Graph Neural Networks (EGNNs); these EGNNs are then used in Sections \ref{sec:number} - \ref{sec:edge_and_properties} to propose forms for each of the subdistributions with the appropriate invariance properties.  The main result appears in Section \ref{sec:vertex}, which shows that the vertex subdistribution can be implemented as a particular type of continuous normalizing flow.

\subsection{Notation and Goal}
\label{sec:notation}

\parbasic{Notation} We use the following notation.  A molecular graph is given by
\begin{equation}
    \mathcal{G} = (N, V, E, A)
\end{equation}
where $N$ is the number of atoms in the molecule; $V$ is the list of vertices, which are the atoms; $E$ is the list of edges, which are the bonds; and $A$ is the set of global molecular properties, i.e. properties which apply to the entire molecule.  The vertex list\footnote{We use lists, rather than sets, so as to make the action of permutations clear.} is $V = \left( \mathbf{v}_i \right)_{i=1}^N$ where each vertex is specified by a vector $\mathbf{v}_i = (\mathbf{x}_i, \mathbf{h}_i)$; in which $\mathbf{x}_i \in \mathbb{R}^3$ is the position of the atom, and $\mathbf{h}_i \in \mathbb{R}^{d_h}$ contains the properties of the atom, such as the atom type.  More generally, this may include continuous properties, discrete ordinal properties, and discrete categorical properties (using the one-hot representation); $\mathbf{h}_i$ may be thought of a concatenation of all such properties.  The edges in the graph $\mathcal{G}$ are undirected, and the edge list $E$ is specified by a neighbourhood relationship.  Specifically, if $\eta_i$ is the set of vertex $i$'s neighbours, then we write $E = \left( \mathbf{e}_{ij} \right)_{i<j: j \in \eta_i}$.  The vector $\mathbf{e}_{ij} \in \mathbb{R}^{d_e}$ contains the properties of the bond connecting atom $i$ and atom $j$, such as the bond type; more generally, $\mathbf{e}_{ij}$ may contain a concatenation of various properties in a manner analogous to the atom properties $\mathbf{h}_i$ as described above.  Finally, the graph properties are given by $K$ individual properties, i.e. $A = (\mathbf{a}_1, \dots, \mathbf{a}_K)$.  A given property $\mathbf{a}_k$ can be either continuous, categorical or ordinal. 

\parbasic{Rigid Transformations}  The action\footnote{
Note to the reader: other papers such as \cite{satorras2021flows} use the notation $E(n)$ rather than $E(3)$, as the transformation is applied to multiple atoms.  Here, we opt to use the notation $E(3)$, and $O(3)$ in the case of rotations, as there is a single transformation being applied to all of the atoms.  Our notation is made sensible and precise given the definitions in Equation (\ref{eq:rigid_action}) and the surrounding text.
}
of a rigid transformation $T \in E(3)$ on a graph $\mathcal{G}$ is given by $T\mathcal{G} = (TN, TV, TE, TA)$ where
\begin{equation}
    TV = \left( T\mathbf{v}_i \right)_{i=1}^N \quad \text{with} \quad T\mathbf{v}_i = (T\mathbf{x}_i, \mathbf{h}_i)
    \label{eq:rigid_action}
\end{equation}
and the other variables are unaffected by $T$; that is $TN = N$, $TE = E$, and $TA = A$.

\parbasic{Permutations} The action of a permutation $\pi \in \mathbb{S}_N$ on a graph $\mathcal{G}$ with $N(\mathcal{G}) = N$ is given by $\pi \mathcal{G} = (\pi N, \pi V, \pi E, \pi A)$ where
\begin{equation}
    \pi V = \left( \mathbf{v}_{\pi_i} \right)_{i=1}^N \quad\quad \text{and} \quad\quad
    \pi E = \left( \mathbf{e}_{\pi_i\pi_j} \right)_{i<j: j \in \eta_i}
    \label{eq:permutation_action}
\end{equation}
and the other variables are unaffected by $\pi$; that is, $\pi N = N$ and $\pi A = A$.

\parbasic{Goal} We assume that we have both a receptor and a ligand, each of which is specified by a molecular graph.  We denote
\begin{equation}
    \mathcal{G} = (N, V, E, A) = \text{ligand graph} \quad \quad \text{and} \quad \quad \hat{\mathcal{G}} = (\hat{N}, \hat{V}, \hat{E}, \hat{A}) = \text{receptor graph}
\end{equation}
(Note: if there molecular properties of the entire ligand-receptor complex, these are subsumed into the ligand molecular properties, $A$.)  Our goal is to learn a conditional generative model: given the receptor, we would like to generate possible ligands.  Formally, we want to learn
\begin{equation}
    p(\mathcal{G} | \hat{\mathcal{G}})
\end{equation}
We want our generative model to observe two types of symmetries.  First, if we transform both the ligand and the receptor with the same rigid transformation, the probability should not change:
\begin{equation}
    p(T\mathcal{G} | T\hat{\mathcal{G}}) = p(\mathcal{G} | \hat{\mathcal{G}}) \quad \text{for all } T \in E(3)
    \label{eq:invariance_rigid}
\end{equation}
Second, permuting the order of either the ligand or the receptor should not affect the probability:
\begin{equation}
    p(\pi \mathcal{G} | \hat{\pi} \hat{\mathcal{G}}) = p(\mathcal{G} | \hat{\mathcal{G}}) \quad \text{for all } \pi \in \mathbb{S}_{N}, \, \hat{\pi} \in \mathbb{S}_{\hat{N}}
    \label{eq:invariance_permutation}
\end{equation}

\subsection{Decomposition of the Conditional Distribution and Invariance Properties}
\label{sec:decomposition}

\parbasic{Decomposition} We may breakdown the conditional generative model as follows:
\begin{equation}
    p(\mathcal{G} | \hat{\mathcal{G}}) = p(N, V, E, A | \hat{\mathcal{G}}) = p(N | \hat{\mathcal{G}}) \, \cdot \, p(V | N, \hat{\mathcal{G}}) \, \cdot \, p(E| N, V, \hat{\mathcal{G}}) \, \cdot \, p(A | N, V, E, \hat{\mathcal{G}})
    \label{eq:markov}
\end{equation}
We refer the four terms on the right-hand side of the equation as the Number Distribution, the Vertex Distribution, the Edge Distribution, and the Property Distribution, respectively.  We will specify a model for each of these distributions in turn.  First, however, we examine how the invariance properties affect the distributions.  

\parbasic{Invariance Properties} Using the relations for rigid body transformations in (\ref{eq:rigid_action}), we have that 
\begin{equation}
    p(T\mathcal{G} | T\hat{\mathcal{G}}) = p(N | T\hat{\mathcal{G}}) \, \cdot \, p(TV | N, T\hat{\mathcal{G}}) \, \cdot \, p(E| N, TV, T\hat{\mathcal{G}}) \, \cdot \, p(A | N, TV, E, T\hat{\mathcal{G}})
    \label{eq:markov_with_rigid}
\end{equation}
Similarly, using the relations for permutations in (\ref{eq:permutation_action}), we have that
\begin{align}
    p(\pi \mathcal{G} | \hat{\pi} \hat{\mathcal{G}}) = p(N | \hat{\pi} \hat{\mathcal{G}}) \, \cdot \, p(\pi V | N, \hat{\pi} \hat{\mathcal{G}}) \, \cdot \, p(\pi E| N, \pi V, \hat{\pi} \hat{\mathcal{G}}) \, \cdot \, p(A | N, \pi V, \pi E, \hat{\pi} \hat{\mathcal{G}}) 
    \label{eq:markov_with_permutation}
\end{align}

Comparing Equation (\ref{eq:markov}) with (\ref{eq:markov_with_rigid}) and (\ref{eq:markov_with_permutation}), the following conditions are sufficient for conditional rigid body invariance (\ref{eq:invariance_rigid}) and conditional permutation invariance (\ref{eq:invariance_permutation}):
\begin{align}
    & p(N | T\hat{\mathcal{G}}) = p(N | \hat{\mathcal{G}})
    & & p(N | \hat{\pi} \hat{\mathcal{G}}) = p(N | \hat{\mathcal{G}}) \label{eq:number_invariance} \\
    & p(TV | N, T\hat{\mathcal{G}}) = p(V | N, \hat{\mathcal{G}}) 
    & & p(\pi V | N, \hat{\pi} \hat{\mathcal{G}}) = p(V | N, \hat{\mathcal{G}}) \label{eq:vertex_invariance} \\
    & p(E| N, TV, T\hat{\mathcal{G}}) = p(E| N, V, \hat{\mathcal{G}}) 
    & & p(\pi E| N, \pi V, \hat{\pi} \hat{\mathcal{G}}) = p(E| N, V, \hat{\mathcal{G}}) \label{eq:edge_invariance} \\
    & p(A | N, TV, E, T\hat{\mathcal{G}}) = p(A | N, V, E, \hat{\mathcal{G}}) \quad
    & & p(A | N, \pi V, \pi E, \hat{\pi} \hat{\mathcal{G}}) = p(A | N, V, E, \hat{\mathcal{G}}) \label{eq:property_invariance}
\end{align}

\subsection{Intermezzo: Two Flavours of EGNNs}
\label{sec:intermezzo}

In order to incorporate the relevant invariance properties, it will be helpful to use Equivariant Graph Neural Networks, also known as EGNNs \citep{satorras2021egnn}.  We now introduce two separate flavours of EGNNs, one which applies to the receptor alone, and a second which applies to the combination of the receptor and the ligand.

\parbasic{Receptor EGNN} \label{sec:receptor_egnn}
This is the standard EGNN which is described in \citep{satorras2021egnn}, applied to the receptor.  As we are dealing with the receptor we use hatted variables:
\begin{align}
    & \bh{m}^\ell_{ij} = \hat{\phi}_e(\bh{h}^\ell_i, \bh{h}^\ell_j, \|\bh{x}^\ell_i-\bh{x}^\ell_j\|^2, \|\bh{x}^0_i-\bh{x}^0_j\|^2, \bh{e}_{ij}, \{\bh{a}_k\})
    \hspace{0.65cm}
    \hat{b}_{ij}^\ell = \sigma(\hat{\phi}_b(\bh{m}_{ij}^\ell))
    \hspace{0.55cm}
    \bh{m}_i^\ell  = \sum_{j \in \hat{\eta}_i} \hat{b}_{ij}^\ell \bh{m}_{ij}^\ell
    \notag \\
    & \bh{x}^{\ell+1}_i=\bh{x}^\ell_i+\left(\sum_{j\neq i} {\frac{(\bh{x}^\ell_i-\bh{x}^\ell_j)}{\|\bh{x}^\ell_i-\bh{x}^\ell_j\|+1}} \right) \hat{\phi}_x(\bh{m}^\ell_{ij})
    \hspace{1.7cm}
    \bh{h}^{\ell+1}_i=\bh{h}^{\ell}_i + \hat{\phi}_h(\bh{h}^\ell_i,\bh{m}^\ell_i)
\end{align}
The particular Receptor EGNN is thus specified by the functions $\hat{\phi}_e, \hat{\phi}_b, \hat{\phi}_x, \hat{\phi}_h$. In practice, prior to applying the EGNN one may apply: (i) an ActNorm layer to the atom positions and features, as well as the edge features and graph features; (ii) a linear transformation to the atom properties.

\parbasic{Receptor-Conditional Ligand EGNN}
It is possible to design a joint EGNN on the receptor and the ligand, by constructing a single graph to capture both.  The main problem with this approach is that the receptor is much larger (1-2 orders of magnitude) than the ligand.  As a result, this naive approach will lead to a situation in which the ligand is ``drowned out'' by the receptor, making it difficult to learn about the ligand.

We therefore take a different approach: we compute summary signatures of the receptor based on the Receptor EGNN, and use these as input to a ligand EGNN.  Our signatures will be based on the feature variables $\bh{h}^\ell_j$ from each layer $\ell = 1, \dots, \hat{L}$.  These variables are invariant to rigid body transformations by construction; furthermore, we can introduce permutation invariance by averaging, that is $\bh{h}^\ell_{av} = \frac{1}{\hat{N}} \sum_{j=1}^{\hat{N}} \bh{h}^\ell_j$.
In Section \ref{sec:vertex}, we will see that the vertex distribution is described by a continuous normalizing flow.  In anticipation of this, we wish to introduce a dependence (crucial in practice) on the time variable $t$ of the ODE corresponding to this flow.  Thus, the receptor at layer $\ell$ of the ligand's EGNN is summarized by the signature $\bh{g}^\ell$ which depends on both $\{ \bh{h}^\ell_{av} \}$ and $t$:
\begin{equation}
    \bh{g}^0 = \phi^0_g \left( \bh{h}^1_{av}, \dots, \bh{h}^{\hat{L}}_{av}, t \right) \qquad \text{and} \qquad
    \bh{g}^\ell = \phi^\ell_g \left( \bh{g}^{\ell-1} \right) \quad \ell = 1, \dots, L
\end{equation}

These invariant receptor signatures $\{ \bh{g}^\ell \}_{\ell=1}^L$ are then naturally incorporated into the Receptor-Conditional Ligand EGNN as follows:
\begin{align}
    & \mathbf{m}^\ell_{ij} = \phi_e\left(\mathbf{h}^\ell_i,\mathbf{h}^\ell_j,  \|\mathbf{x}^\ell_i-\mathbf{x}^\ell_j\|^2, \|\mathbf{x}^0_i-\mathbf{x}^0_j\|^2 , \bh{g}^\ell \right)
    \hspace{0.5cm}
    \quad b_{ij}^\ell = \sigma(\phi_a(\mathbf{m}_{ij}^\ell, \bh{g}^\ell ))
    \hspace{0.5cm}
    \mathbf{m}^\ell_i  = \sum_{j=1}^N b^\ell_{ij} \mathbf{m}^\ell_{ij}
    \notag \\
    & \mathbf{x}^{\ell+1}_i= \mathbf{x}^\ell_i + \left( \sum_{j\neq i} {\frac{(\mathbf{x}^\ell_i-\mathbf{x}^\ell_j)}{\|\mathbf{x}^\ell_i-\mathbf{x}^\ell_j\|+1}} \right) \phi_x(\mathbf{m}^\ell_{ij} , \bh{g}^\ell )
    \hspace{0.7cm}
    \mathbf{h}^{\ell+1}_i = \mathbf{h}^{\ell}_i + \phi_h(\mathbf{h}^\ell_i, \mathbf{m}^\ell_i , \bh{g}^\ell )
\end{align}
The particular Receptor-Conditional Ligand EGNN is thus specified by the functions $\{ \phi^\ell_g \}_{\ell=0}^L, \phi_e, \phi_b, \phi_x, \phi_h$.

\subsection{The Number Distribution: $p(N | \hat{\mathcal{G}})$}
\label{sec:number}

\parbasic{Construction} Given the invariance conditions for the number distribution $p(N | \hat{\mathcal{G}})$ described in Equation (\ref{eq:number_invariance}), we propose the following distribution.  Let $\boldsymbol{\zeta}_N$ indicate a one-hot vector, where the index corresponding to $N$ is filled in with a $1$.  Based on the output of the receptor EGNN, compute
\begin{equation}
    p(N | \hat{\mathcal{G}}) = \boldsymbol{\zeta}_N^T \texttt{MLP}' \left( \frac{1}{\hat{N}} \sum_{i=1}^{\hat{N}} \texttt{MLP} \left( \bh{h}^L_i \right) \right)
\end{equation}
where the outer $\texttt{MLP}$'s last layer is a softmax of size equal to the maximum number of atoms allowed.

Due to the fact that we use the $\bh{h}^L_i$ vectors (and not the $\bh{x}^L_i$ vectors), we have the first invariance condition, as $T\bh{h}^L_i = \bh{h}^L_i$.  Due to the fact that we use an average, we have the second invariance condition.  Note that we can choose to make the inner $\texttt{MLP}$ the identity, if we so desire.

\parbasic{Loss Function}  The loss function is straightforward here -- it is simply the negative log-likelihood of the number distribution, i.e. $L(\theta) = \mathbb{E}_{\mathcal{G}, \hat{\mathcal{G}}} \left[ -\log p(N | \hat{\mathcal{G}}; \theta) \right]$.

\subsection{The Vertex Distribution: $p(V | N, \hat{\mathcal{G}})$ via Continuous Normalizing Flows}
\label{sec:vertex}

\parbasic{General Notation} Given the invariance conditions for the vertex distribution $p(V | N, \hat{\mathcal{G}})$ described in Equation (\ref{eq:vertex_invariance}), we now outline a procedure for constructing such a distribution.  We begin with some notation.  Let us define a vectorization operation on the vertex list $V$, which produces a vector $\mathbf{v}$; we refer to this as a \textit{vertex vector}.  Recall that $V = (\mathbf{v}_i)_{i=1}^N$ where $\mathbf{v}_i = (\mathbf{x}_i, \mathbf{h}_i)$.  Let
\begin{equation}
    \mathbf{x} = \texttt{concat}(\mathbf{x}_1, \dots, \mathbf{x}_N) 
    \hspace{0.8cm}
    \mathbf{h} = \texttt{concat}(\mathbf{h}_1, \dots, \mathbf{h}_N)
    \hspace{0.8cm}
    \mathbf{v} = \texttt{concat}(\mathbf{x}, \mathbf{h})
\end{equation}
The vertex vector $\mathbf{v} \in \mathbb{R}^{d_v^N}$ where $d_v^N = (d_h+3)N$. We denote the mapping from the vertex list $V$ to the vertex vector $\mathbf{v}$ as the vectorization operation $\texttt{vec}(\cdot)$:
\begin{equation}
    \mathbf{v} = \texttt{vec}(V) \quad \quad \text{ and } \quad \quad V = \texttt{vec}^{-1}(\mathbf{v})
\end{equation}
We have already described the action of rigid body transformation $T$ and permutations $\pi$ on the vertex list $V$ in Equations (\ref{eq:rigid_action}) and (\ref{eq:permutation_action}).  It is easy to extend this to vertex vectors $\mathbf{v}$ using the \texttt{vec} operation; we have $T\mathbf{v} = \texttt{vec}( T\texttt{vec}^{-1}(\mathbf{v}) )$ and $\pi\mathbf{v} = \texttt{vec}( \pi\texttt{vec}^{-1}(\mathbf{v}) )$.

Given the above, it is sufficient for us to describe the distribution $p_{vec}(\mathbf{v} | \hat{\mathcal{G}})$
from which the vertex distribution $p(V | N, \hat{\mathcal{G}})$ follows directly, $p(V | N, \hat{\mathcal{G}}) = p_{vec}(\texttt{vec}(V) | \hat{\mathcal{G}})$.  Note that we have suppressed $N$ in the condition in $p_{vec}(\cdot)$, as $\mathbf{v}$ is a vector of dimension $d_v^N$, so the $N$ dependence is already implicitly encoded.  

\parbasic{Complex-to-Ligand Mapping and Semi-Equivariance} Let $\gamma$ be a function which takes as input the ligand graph $\mathcal{G}$ and receptor graph $\hat{\mathcal{G}}$, and outputs a new vertex list $V'$ for the ligand graph $\mathcal{G}$:
\begin{equation}
    \gamma: \mathcal{\mathcal{G}} \times \hat{\mathcal{\mathcal{G}}} \to \mathcal{V} \quad \quad V' = \gamma(\mathcal{G}, \hat{\mathcal{G}})
\end{equation}
We refer to $\gamma$ as a \textit{Complex-to-Ligand Mapping}.  A rigid body transformation $T \in E(3)$ consists of a rotation and translation; let the rotation be denoted as $T_{rot}$.  Then we say that $\gamma$ is \textit{rotation semi-equivariant} if
\begin{equation}
    \gamma(T_{rot}\mathcal{G}, T\hat{\mathcal{G}}) = T_{rot}\gamma(\mathcal{G}, \hat{\mathcal{G}}) \quad \text{for all } T \in E(3)
\end{equation}
where, as before, the action of $T$ on a graph is given by Equation (\ref{eq:rigid_action}).  $\gamma$ is said to be \textit{permutation semi-equivariant} if
\begin{equation}
    \gamma(\pi \mathcal{G}, \hat{\pi}\hat{\mathcal{G}}) = \pi \gamma(\mathcal{G}, \hat{\mathcal{G}}) \quad \text{for all } \pi \in \mathbb{S}_N \text{ and } \hat{\pi} \in \mathbb{S}_{\hat{N}}
    \label{eq:gamma_permutation}
\end{equation}
where, as before, the action of the permutation on a graph is given by Equation (\ref{eq:permutation_action}).  Note in the definitions of both types of semi-equivariance, the differing roles played by the ligand and receptor; as the equivariant behaviour only applies to the ligand, we have used the term semi-equivariance.



\parbasic{Receptor-Conditioned Ligand Flow} Let $\gamma: \mathcal{\mathcal{G}} \times \hat{\mathcal{\mathcal{G}}} \to \mathcal{V}$ be a Complex-to-Ligand Mapping.  If $\mathbf{v}$ is a vertex vector, define $\mathcal{G}_{\mathbf{v}}$ to be the graph $\mathcal{G}$ with the vertex set replaced by $\texttt{vec}^{-1}(\mathbf{v})$. Then the following ordinary differential equation is referred to as a \textit{Receptor-Conditioned Ligand Flow}:
\begin{equation}
    \frac{d\mathbf{u}}{dt} = \texttt{vec}\left( \gamma(\mathcal{G}_{\mathbf{u}} , \hat{\mathcal{G}}) \right), \quad \text{with } \mathbf{u}(0) = \mathbf{z}
\end{equation}
where the initial condition $\mathbf{z} \sim \mathcal{N}(0, \mathbf{I})$ is a Gaussian random vector of dimension $d_v^N$, and the ODE is run until $t=1$.  $\mathbf{u}(1)$ is thus the output of the Receptor-Conditioned Ligand Flow.

\parbasic{Vertex Distributions with Appropriate Invariance} We now have all of the necessary ingredients in order to construct a distribution $p_{vec}(\mathbf{v} | \hat{\mathcal{G}})$ which yields a vertex distribution $p(V | N, \hat{\mathcal{G}})$ that satisfies the invariance conditions that we require.  The following is our main result:

\begin{theorem}
Let $\mathbf{u}(1)$ be the output of a Receptor-Conditioned Ligand Flow specified by the Complex-to-Ligand Mapping $\gamma$.  Let the mean position of the receptor be given by $\bh{x}_{av} = \frac{1}{\hat{N}} \sum_{i=1}^{\hat{N}} \bh{x}_i$, and define the following quantities
\begin{equation}
    \alpha = \frac{N}{N + \hat{N}} \quad \quad
    \Omega_{\hat{\mathcal{G}}} = 
    \begin{bmatrix}
        \mathbf{I}_{3N} - \frac{\alpha}{N} \mathbf{1}_{N \times N} \otimes \mathbf{I}_3 & \mathbf{0} \\
        \mathbf{0} & \mathbf{I}_{d_h N}
    \end{bmatrix}
    \quad \quad
    \omega_{\hat{\mathcal{G}}} = 
    \begin{bmatrix}
        -(1-\alpha) \mathbf{1}_{N \times 1} \otimes \bh{x}_{av} \\
        \mathbf{0}
    \end{bmatrix}
\end{equation}
where $\otimes$ indicates the Kronecker product.  Finally, let 
\begin{equation}
    \mathbf{v} = \Omega_{\hat{\mathcal{G}}}^{-1} \left( \mathbf{u}(1) - \omega_{\hat{\mathcal{G}}} \right)
\end{equation}

Suppose that $\gamma$ is both rotation semi-equivariant and permutation semi-equivariant.  Then the resulting distribution on $\mathbf{v}$, that is $p_{vec}(\mathbf{v} | \hat{\mathcal{G}})$, yields a vertex distribution $p(V | N, \hat{\mathcal{G}}) = p_{vec}(\texttt{vec}(V) | \hat{\mathcal{G}})$ that satisfies the invariance conditions in Equation (\ref{eq:vertex_invariance}).
\end{theorem}

\textbf{\textit{Proof:}} See Appendix \ref{appnx:proof}.

\parbasic{Designing the Complex-to-Ligand Mapping} Examining the theorem, we see that the one degree of freedom that we have is the Complex-to-Ligand Mapping $\gamma$.  For this, we choose to use the Receptor-Conditional Ligand EGNN, where the output of the Complex-to-Ligand Mapping $V' = \gamma(\mathcal{G}, \hat{\mathcal{G}})$ is simply the final layer, i.e. the $i^{th}$ vertex of $V'$ is given by $v'_i = (\mathbf{x}_i^L, \mathbf{h}_i^L)$.

The rotation semi-equivariance of $\gamma$ follows straightforwardly from the rotation semi-equivariance of EGNNs and the rotation invariance of the receptor signatures $\{ \bh{g}^\ell \}_{\ell=1}^L$.  Similarly, the permutation semi-equivariance follows from the permutation semi-equivariance of EGNNs and the permutation invariance of the receptor signatures.

\parbasic{Loss Function}  The vertex distribution is described by a continuous normalizing flow.  The loss function and its optimization are implemented using standard techniques from this field \citep{chen2018neural,grathwohl2018ffjord,chen2018continuous}.  As the feature vectors can contain discrete variables such as the atom type, then techniques based on variational dequantization \citep{ho2019flow++} and argmax flows \cite{hoogeboom2021argmax} can be used, for ordinal and categorical features respectively.  This is parallel to the treatment in \citep{satorras2021flows}.

\subsection{Extensions: The Edge and Properties Distributions}
\label{sec:edge_and_properties}

We now outline methods for computing the Edge and Properties Distributions.  In practice, we do not implement these methods, but rather use a standard simple technique based on inferring edge properties directly from vertices, see Section \ref{sec:experiments}.  Nevertheless, we describe these methods for completeness.

\parbasic{The Edge Distribution} Given the invariance conditions for the edge distribution $p(E| N, V, \hat{\mathcal{G}})$ described in Equation (\ref{eq:edge_invariance}), we propose
a distribution which displays \textit{conditional independence}: 
$p\left( E = ( \mathbf{e}_{ij} \right)_{i<j: j \in \eta_i} | N, V, \hat{\mathcal{G}} ) = 
    \prod_{i<j: j \in \eta_i} p( \mathbf{e}_{ij} | N, V, \hat{\mathcal{G}} )$.
We opt for conditional independence for two reasons: (1) The usual Markov decomposition of the probability distribution with terms of the form $p( \mathbf{e}_{ij} | \mathbf{e}_{<ij}, N, V, \hat{\mathcal{G}} )$ implies a particular ordering of the edges, and is therefore not permutation-invariant. (2) $V$ is a deterministic and invertible function of the flow's noise vector $\mathbf{z}$; thus, conditioning on $V$ is the same as conditioning on $\mathbf{z}$.  If $E$ is a deterministic (but not necessarily invertible) function of $\mathbf{z}$, then conditional independence is correct.

To compute $p( \mathbf{e}_{ij} | N, V, \hat{\mathcal{G}} )$, we use a second Receptor-Conditional Ligand EGNN.  The key distinction between this network and the Receptor-Conditional Ligand EGNN used in computing the vertex distribution is the initial conditions.  In the case of the vertex distribution, the initial conditions are $\mathbf{x}_i^1 = \mathbf{0}$ and $\mathbf{h}_i^1 = \mathbf{0}$.  In the current case of the edge distribution, we are given $V$ (we are conditioning on it); thus, we take the initial conditions to be $\bt{x}_i^1 = \mathbf{x}_i(V)$ and $\bt{h}_i^1 = \mathbf{h}_i(V)$.  In other words, the initial values are given the vertex list $V$ itself.

Given this second Receptor-Conditional Ligand EGNN, we can compute the edge distribution as
\begin{equation}
    p\left( \mathbf{e}_{ij} | N, V, \hat{\mathcal{G}} \right) = \mathbf{e}_{ij}^T \, \texttt{MLP}\left(\bt{m}^L_{ij}\right)
\end{equation}
in the case of categorical properties (where $\texttt{MLP}$'s output is a softmax with $d_e$ entries); analogous expressions exist for ordinal or continuous properties.  It is straightforward to see that this distribution satisfies the invariance properties in Equation (\ref{eq:edge_invariance}).  The corresponding loss function is a simple cross-entropy loss (or regression loss for non-categorical properties).

\parbasic{The Property Distribution} Given the invariance conditions for the property distribution $p(A | N, V, E, \hat{\mathcal{G}})$ described in Equation (\ref{eq:property_invariance}), we propose the following distribution.  We use a standard Markov decomposition:
$
    p(A | N, V, E, \hat{\mathcal{G}}) = \prod_{k=1}^K p\left( \mathbf{a}_k | \mathbf{a}_{1:(k-1)}, N, V, E, \hat{\mathcal{G}} \right)
$.
Let
\begin{equation}
    \boldsymbol{\xi}_h = \frac{1}{N} \sum_{i=1}^N \texttt{MLP} \left( \bt{h}^L_i \right)
    \hspace{0.8cm}
    \boldsymbol{\xi}_e = \frac{1}{|E|} \sum_{i<j: j \in \eta_i} \texttt{MLP} \left( \mathbf{e}_{ij} \right)
    \hspace{0.8cm}
    \boldsymbol{\xi}_{a,k} = \texttt{MLP}\left( \sum_{j=1}^{k-1} W_j \mathbf{a}_j \right)
\end{equation}
where the matrices $W_1, \dots W_K$ all have the same number of rows.
Then we set
\begin{equation}
    p\left( \mathbf{a}_k \left| \mathbf{a}_{1:(k-1)}, N, V, E, \hat{\mathcal{G}} \right. \right) = \mathbf{a}_k^T \texttt{MLP} \left( \texttt{concat}\left( \boldsymbol{\xi}_h , \boldsymbol{\xi}_e, \boldsymbol{\xi}_{a,k} \right) \right)
\end{equation}
in the case of categorical properties; analogous expressions exist for ordinal or continuous properties.  Note that the only item which changes for the different properties $k$ is the vector $\boldsymbol{\xi}_{a,k}$. It is easy to see that this distribution satisfies the invariance properties in Equation (\ref{eq:property_invariance}).  The corresponding loss function is a simple cross-entropy loss (or regression loss for non-categorical properties).

%% file: text/experiments.tex





\parbasic{Data} We use the CrossDocked2020 dataset \citep{francoeur2020three} which contains poses of ligands docked into multiple similar binding pockets across the Protein Data Bank.  We use the authors' suggested split into training and validation sets.  The dataset contains docked receptor-ligand pairs whose binding pose RMSD is lower than $2\text{\AA}$.  We keep only those data points whose ligand has 30 atoms or fewer with atom types in \{C, N, O, F\}; which do not contain duplicate vertices; and whose predicted Vina scores \citep{trott2010autodock} are within distribution.  The refined datasets consist of 132,863 training data points and 63,929 validation data points.  A full description of the data is summarized in Appendix \ref{appnx:data}.

\parbasic{Features} The ligand features that we wish to predict include the atom type $\in$ \{C, N, O, F\} (categorical); the stereo parity $\in$ \{not stereo, odd, even\} (categorical); and charge $\in \{-1, 0, +1\}$ (ordinal).  The receptor features that are used are computed with the Graphein library \citep{jamasb2022graphein}.  The vertex features include: the atom type $\in$ \{C, N, O, S, ``other''\} where ``other'' is a catch-all for less common atom types\footnote{Specifically: Na, Mg, P, Cl, K, Ca, Co, Cu, Zn, Se, Cd, I, Hg.} (categorical); the Meiler Embeddings \citep{meiler2001generation} (continuous $\in \mathbb{R}^{7}$).  The bond (edge) properties include: the bond order $\in$ \{Single, Double, Triple\} (categorical); covalent bond length (continuous $\in \mathbb{R}$). The receptor overall graph properties ($\hat{A}$) contain the weight of all chains contained within a polypeptide structure, see \citep{jamasb2022graphein}.

\parbasic{Training} Training the model takes approximately 14 days using a single NVIDIA A100 GPU for 30 epochs.  
We train with the Adam optimizer, weight decay of $10^{-12}$, batch size of 128, and learning rate of $2\times10^{-4}$.
We also train the baseline (state of the art) technique, GraphBP \citep{liu2022generating} on our filtered dataset.  We train it for 100 epochs, using the hyperparameters given in the paper, with one exception: we set the atom number range of the autoregressive generative process according to the atom distribution of the filtered dataset.

\input{floats/tab_validity_binding}

\input{floats/fig_bond_dist}

\parbasic{Evaluation} For both the proposed method as well as GraphBP, we perform inference using the method suggested in \cite{liu2022generating}.  Given a receptor, we sample from the learned distribution, which generates the ligands' vertices; we then then apply OpenBabel \citep{hummell2021novel} to construct bonds.  Evaluation follows the standard procedure \citep{francoeur2020three,liu2022generating}.  First, the receptor target is computed by taking all the atoms in the receptor that are less than $15\text{\AA}$ from the center of mass of the reference ligand; if the target has fewer than 200 atoms, the threshold of $15\text{\AA}$ until the 200 atom minimum is reached.  We then generate 100 ligands for each reference binding site in the evaluation set, and compute statistics (i.e. validity and $\Delta$Binding, see below) on this set of samples. As in \citep{francoeur2020three,liu2022generating}, 10 target receptors for evaluation; each target receptor has multiple associated ligands, leading to 90 (receptor, reference-ligand) pairs.

\parbasic{Validity} The validity is defined as the percentage of molecules that are chemically valid among all generated molecules. A molecule is valid if it can be sanitized by RDKit; for an explanation of the sanitization procedure, see \citep{Landrum2016RDKit2016_09_4}. As shown in Table \ref{tab:validity_binding}(a), our model produces ligands with a validity of 99.86\%, surpassing the previous state of the art. We also compute the distribution of bond distances of the two methods, and compare this to distribution of the reference ligands; see Figure \ref{fig:bond_dist}.  Our method's distribution is considerably closer to the reference distribution than GraphBP; some non-trivial fraction of the time, GraphBP produces unusual, very high bond distances.  (In fact, we have discarded values higher than $10\text{\AA}$ on the GraphBP plot so as to display the distributions on similar scales.)  This impression is reinforced in Table \ref{tab:validity_binding}(a) which compares the mean and standard deviation of these distributions.



\parbasic{Binding Affinity} A more interesting measure than validity is $\Delta$Binding, which measures the measures the percentage of generated molecules that have higher predicted binding affinity to the target binding site than the corresponding reference molecule.  To compute binding affinities, we follow the procedure used by GraphBP.  Briefly, we refine the generated 3D molecules by Universal Force Field minimization \citep{rappe1992uff}; then, Vina minimization and CNN scoring are applied to both generated and reference molecules by using gnina, a molecular docking program \citep{mcnutt2021gnina}.  As can be seen in Table \ref{tab:validity_binding}(b), our result improves significantly on the state of the art.  Raw GraphBP attains $\Delta$Binding = 13.45\%.  By playing with the minimum and the maximum atom number of the baseline autoregressive model, we were able to improve this to 22.76\%; however, note that this results in a  reduction in validity from 99.76\% to 99.54\%.  Our method attains $\Delta$Binding = 35.7\%, which is a relative improvement of 56.81\% over the better of the two GraphBP scores.

\parbasic{Qualitative Results} We show examples of generated ligands in Figure \ref{fig:qualitative}, along with their chemical structures.  Note that the structures of the generated molecules differ substantially from the reference molecules, indicating that the model has indeed learn to generalize to interesting novel structures.

\input{floats/fig_qualitative}

%% file: floats/tab_validity_binding.tex

    



\begin{table}[tb]
    \begin{subtable}{.5\linewidth}
      \centering
        \begin{tabular}{ |c||c|c|c|} 
            \hline
            \multicolumn{4}{|c|}{\textbf{Validity}} \\
            \hline
            \multicolumn{2}{|c|}{Ours} & \multicolumn{2}{|c|}{GraphBP} \\
            \hline
            \multicolumn{2}{|c|}{99.87\%} & \multicolumn{2}{|c|}{99.75\%} \\
            \hline
            \hline
            \hline
            \multicolumn{4}{|c|}{\textbf{Bond Length Distribution}} \\
            \hline
            & Ref. Mols. & Ours & GraphBP \\
            \hline
            mean & 1.42 & 1.45 & 1.65 \\
            std & 0.08 & 0.10 & 0.95 \\
            \hline
        \end{tabular}
        \caption{Ligand validity and bond length distribution}
    \end{subtable}%
    \begin{subtable}{.5\linewidth}
      \centering
        \begin{tabular}{ |c||c|c|c|} 
            \hline
            \multicolumn{4}{|c|}{$\mathbf{\Delta}$\textbf{Binding}} \\
            \hline
            \multicolumn{2}{|c|}{Ours} & \multicolumn{2}{|c|}{GraphBP} \\
            \hline
            \multicolumn{2}{|c|}{35.7\%} & \multicolumn{2}{|c|}{22.76\%} \\
            \hline\hline\hline
            \multicolumn{4}{|c|}{\textbf{Predicted Affinity Distribution}} \\
            \hline
            & Ref. Mols. & Ours & GraphBP \\
            \hline
            mean & 5.09 & 4.56 & 4.31 \\
            std & 1.16 & 1.05 & 1.03 \\
            \hline
        \end{tabular}
        \caption{$\Delta$Binding and predicted affinity distribution}
    \end{subtable}
    \caption{Comparison of molecule validity and $\Delta$Binding between proposed method and GraphBP.}
    \label{tab:validity_binding}
\end{table}

%% file: floats/fig_bond_dist.tex
\begin{figure}[tb]

\centering
\includegraphics[width=.25\textwidth]{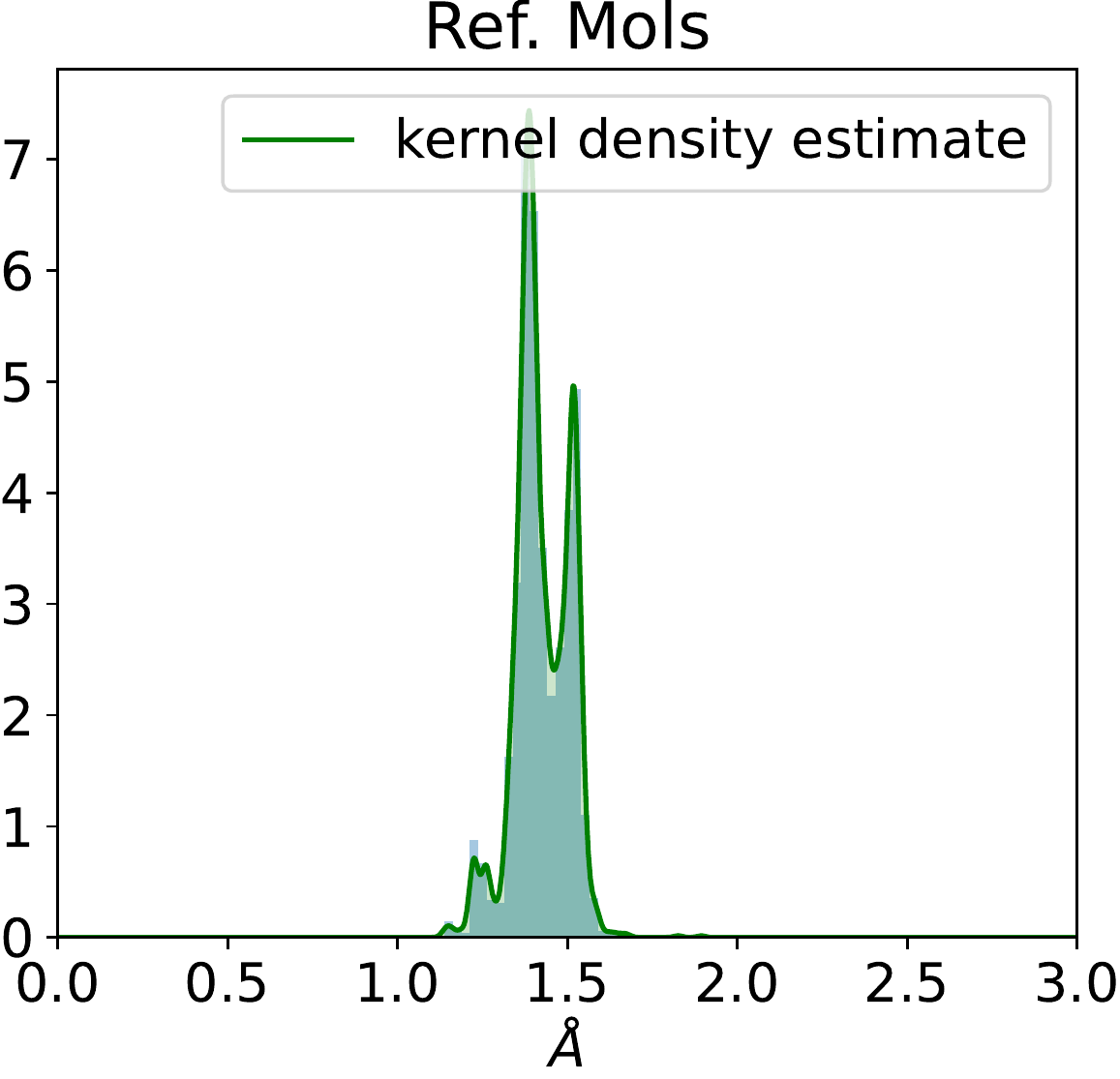}\hfill
\includegraphics[width=.25\textwidth]{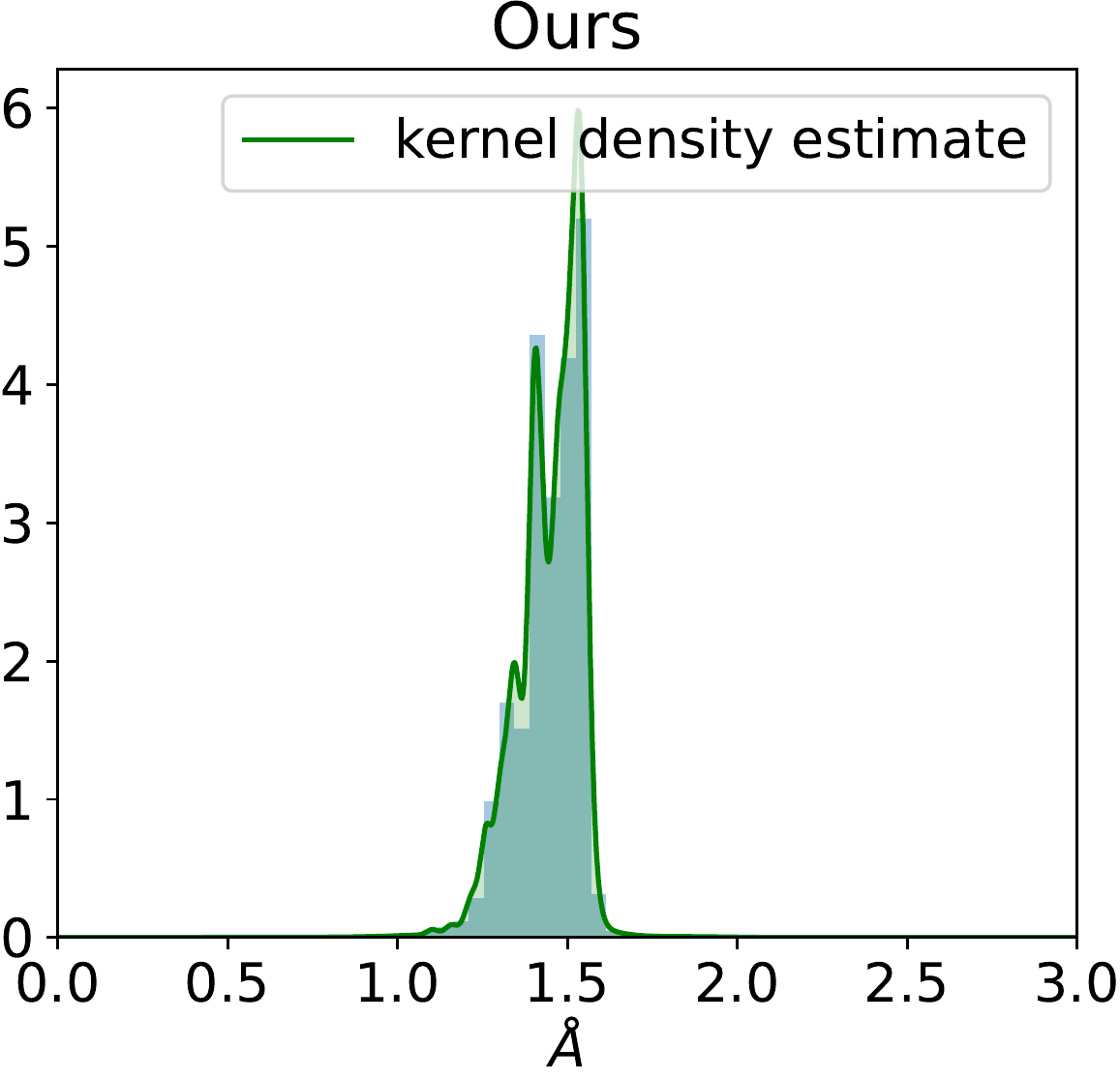}\hfill
\includegraphics[width=.25\textwidth]{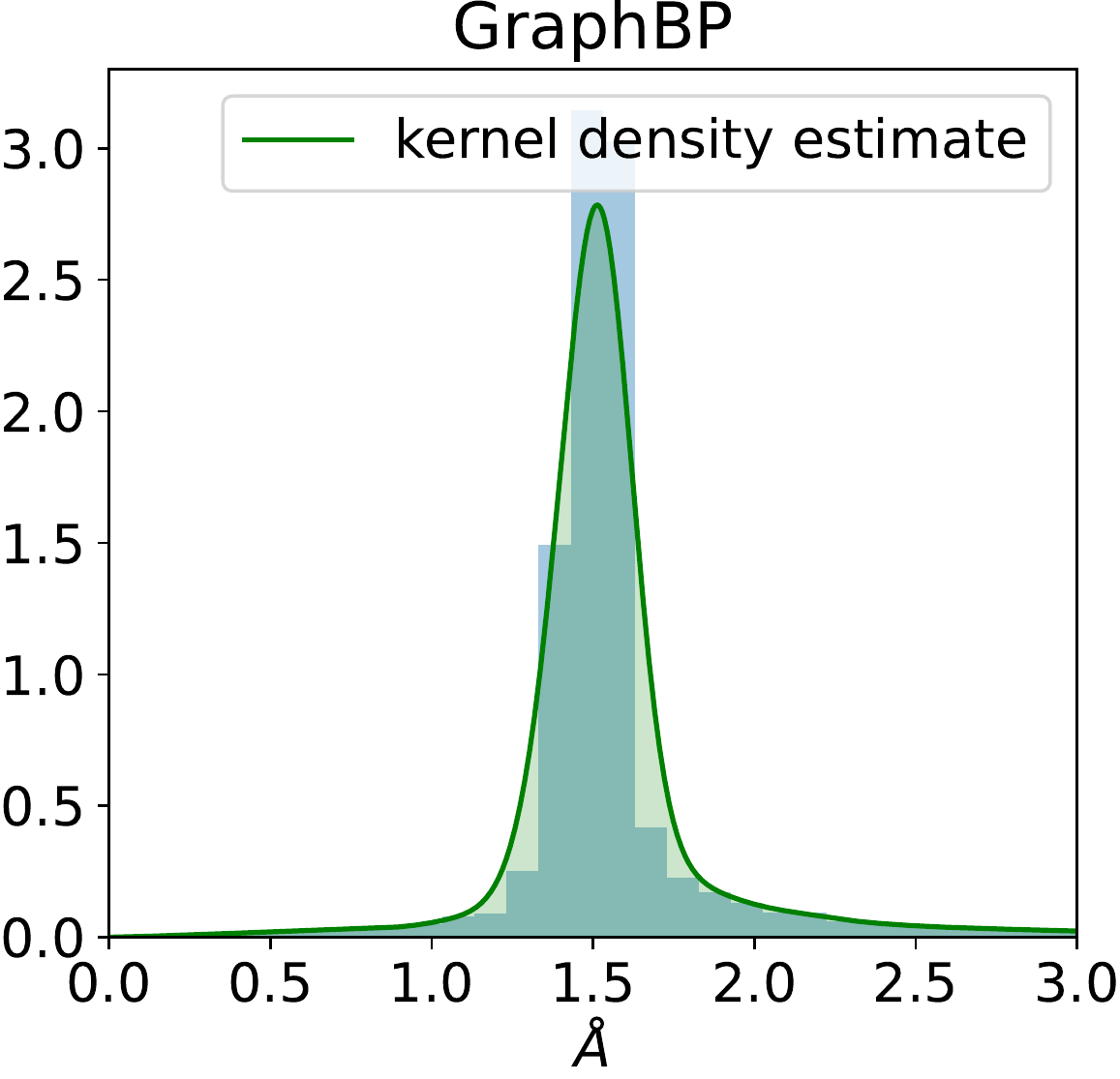}

\caption{Normalized histogram of relative distances between atoms}
\label{fig:bond_dist}

\end{figure}

%% file: floats/fig_qualitative.tex
\begin{figure*}
    \centering
    \begin{tabular}{cccccc}
         \fig{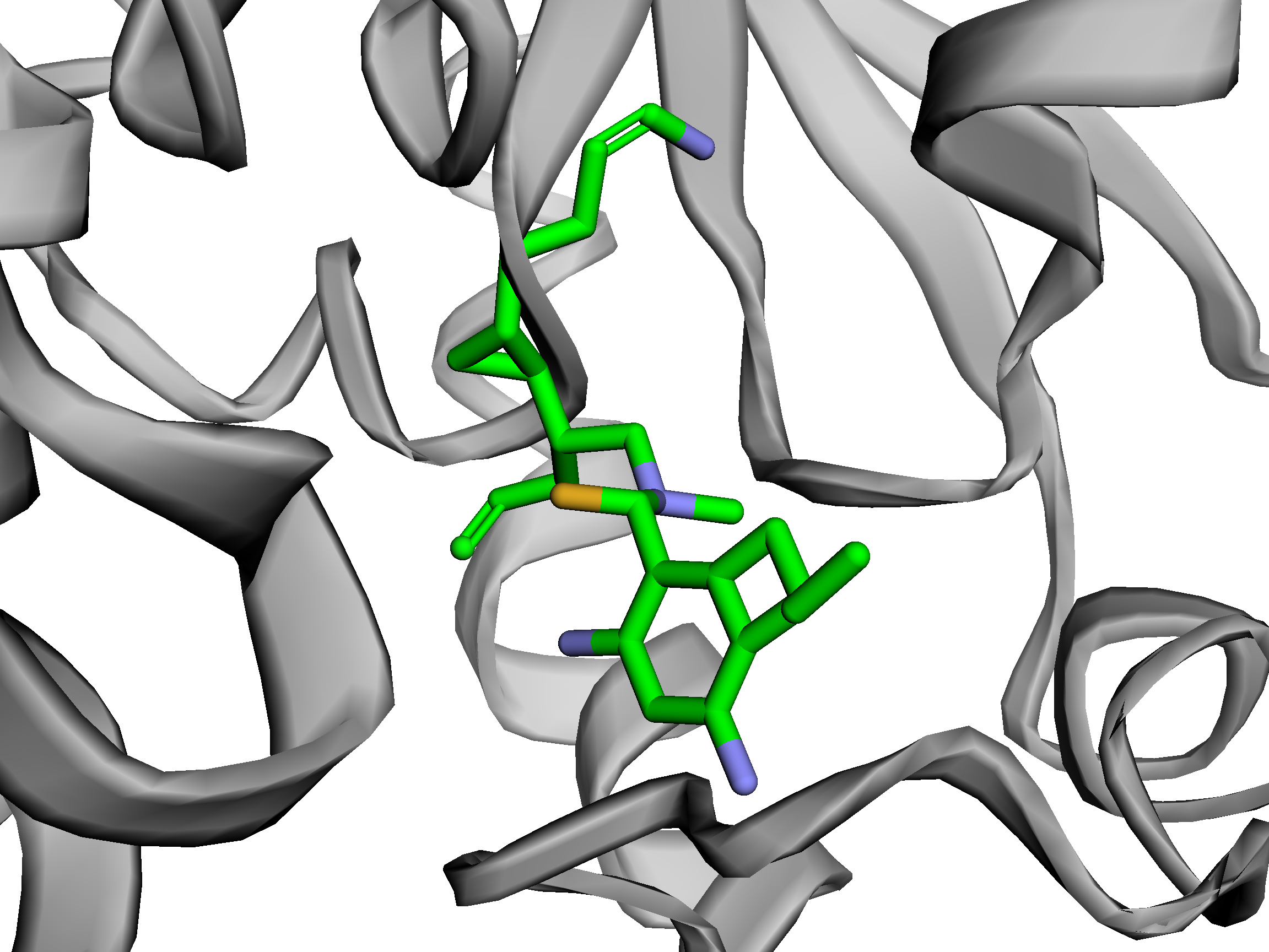}{0.13}
         & \fig{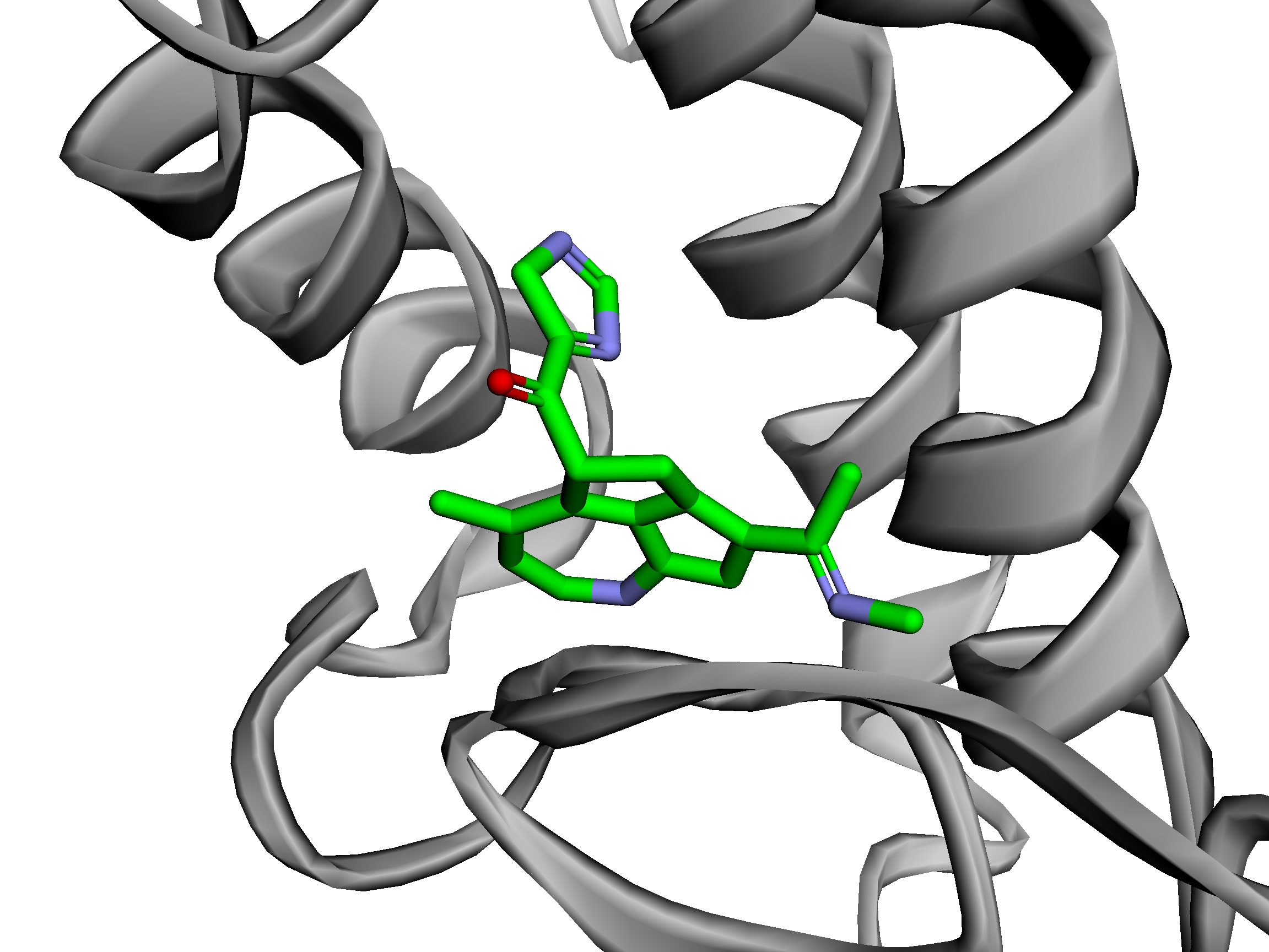}{0.13}
         & \fig{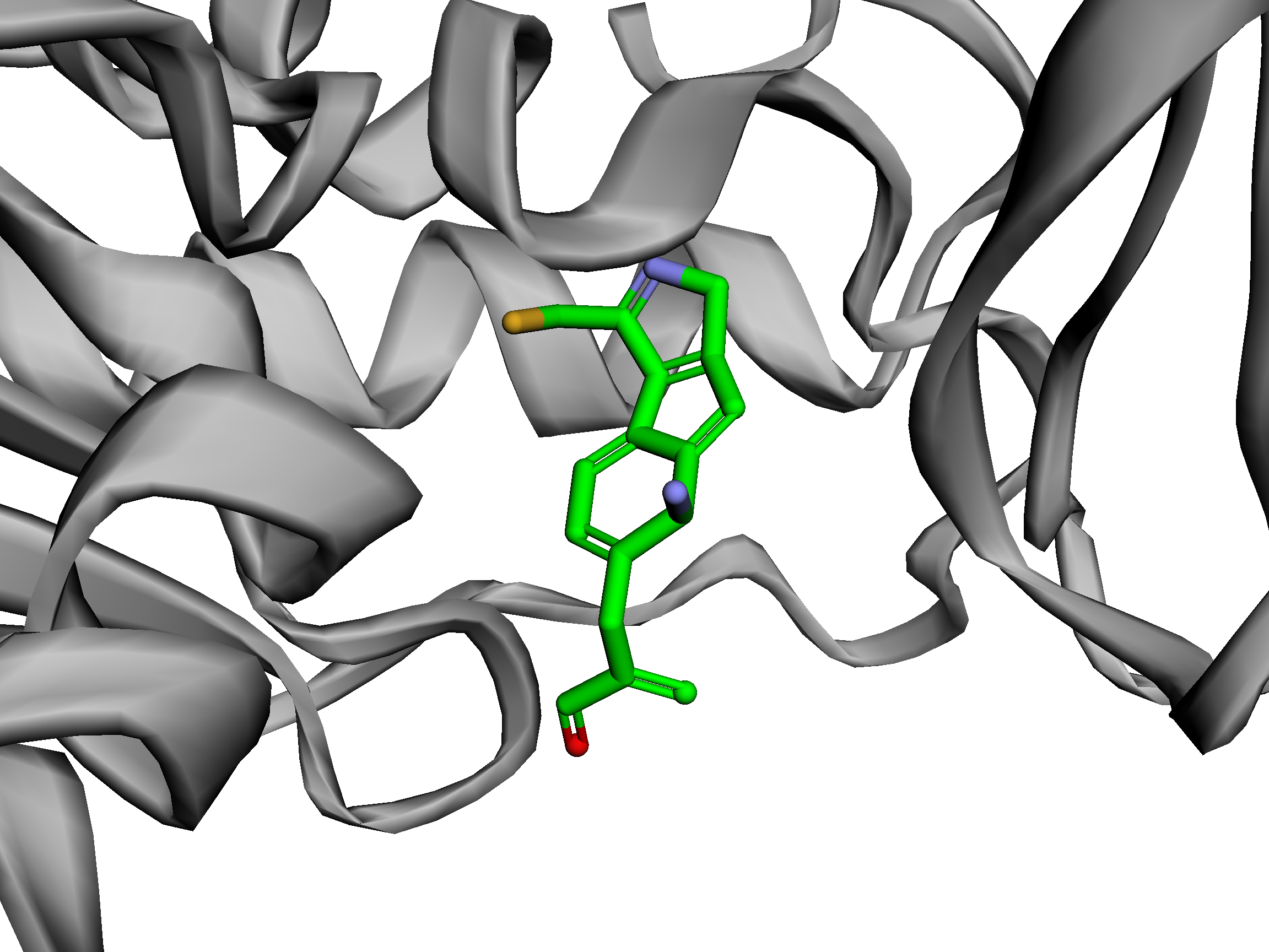}{0.13}
         & \fig{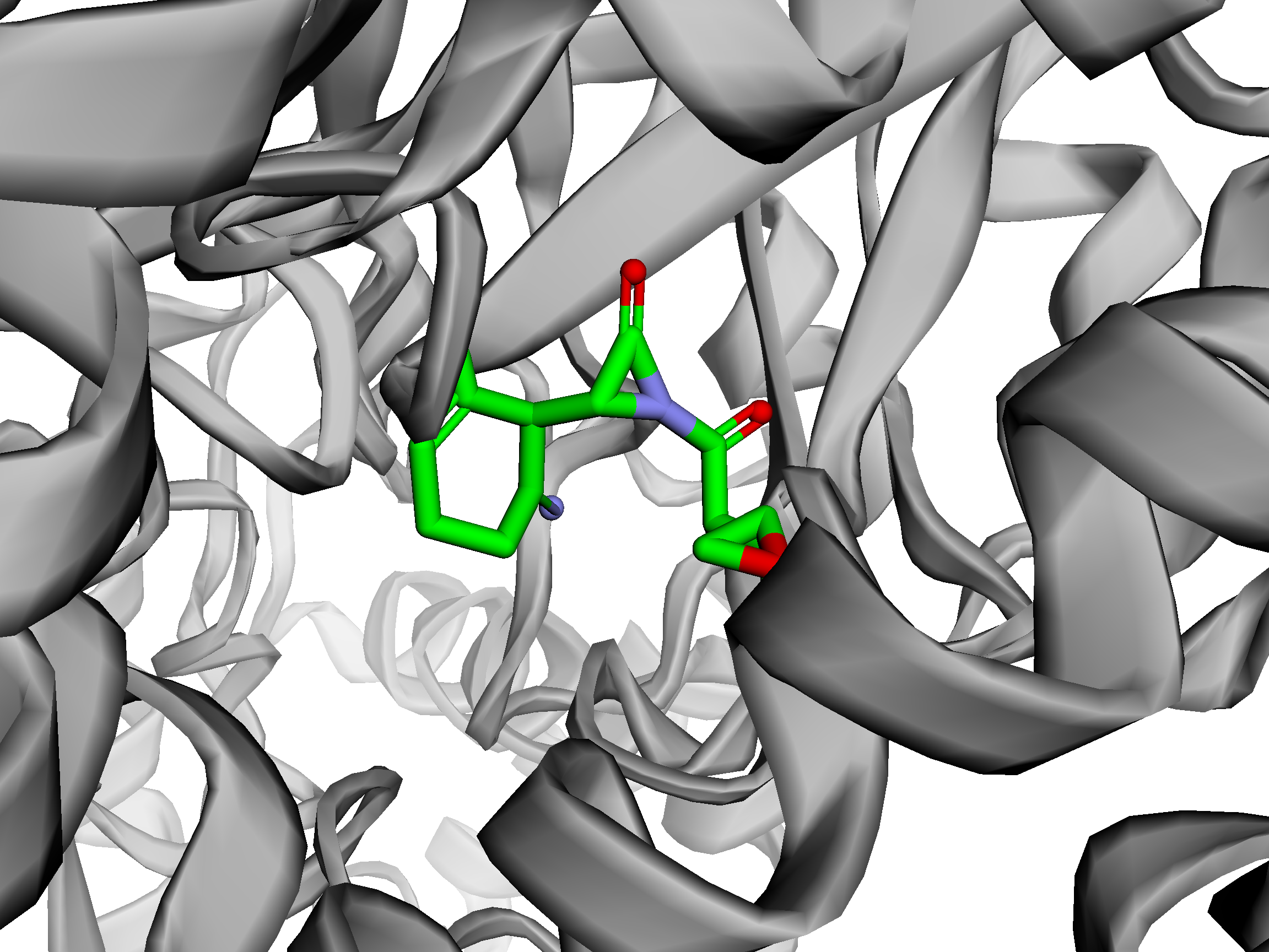}{0.13}
         & \fig{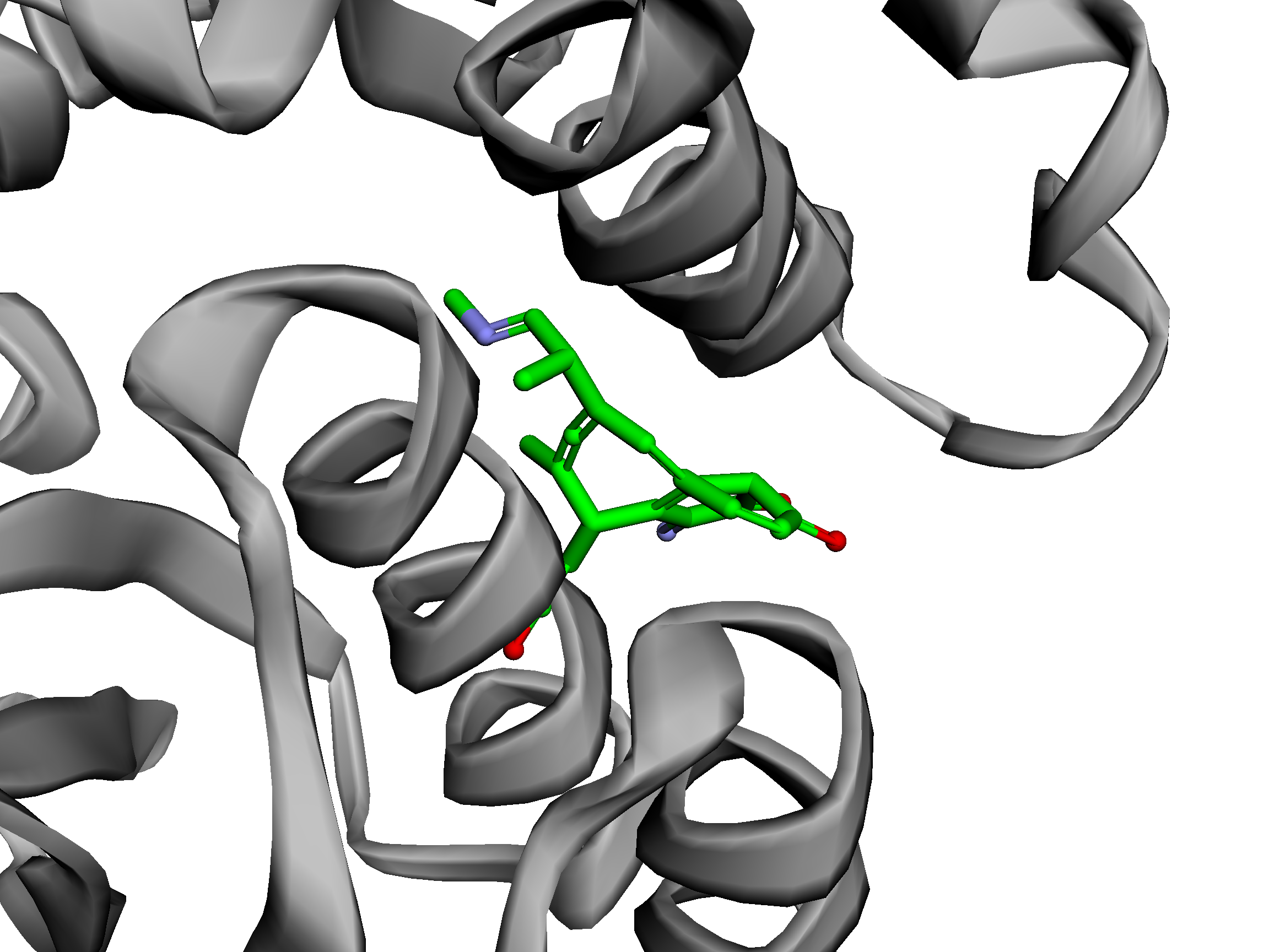}{0.13}
         & \fig{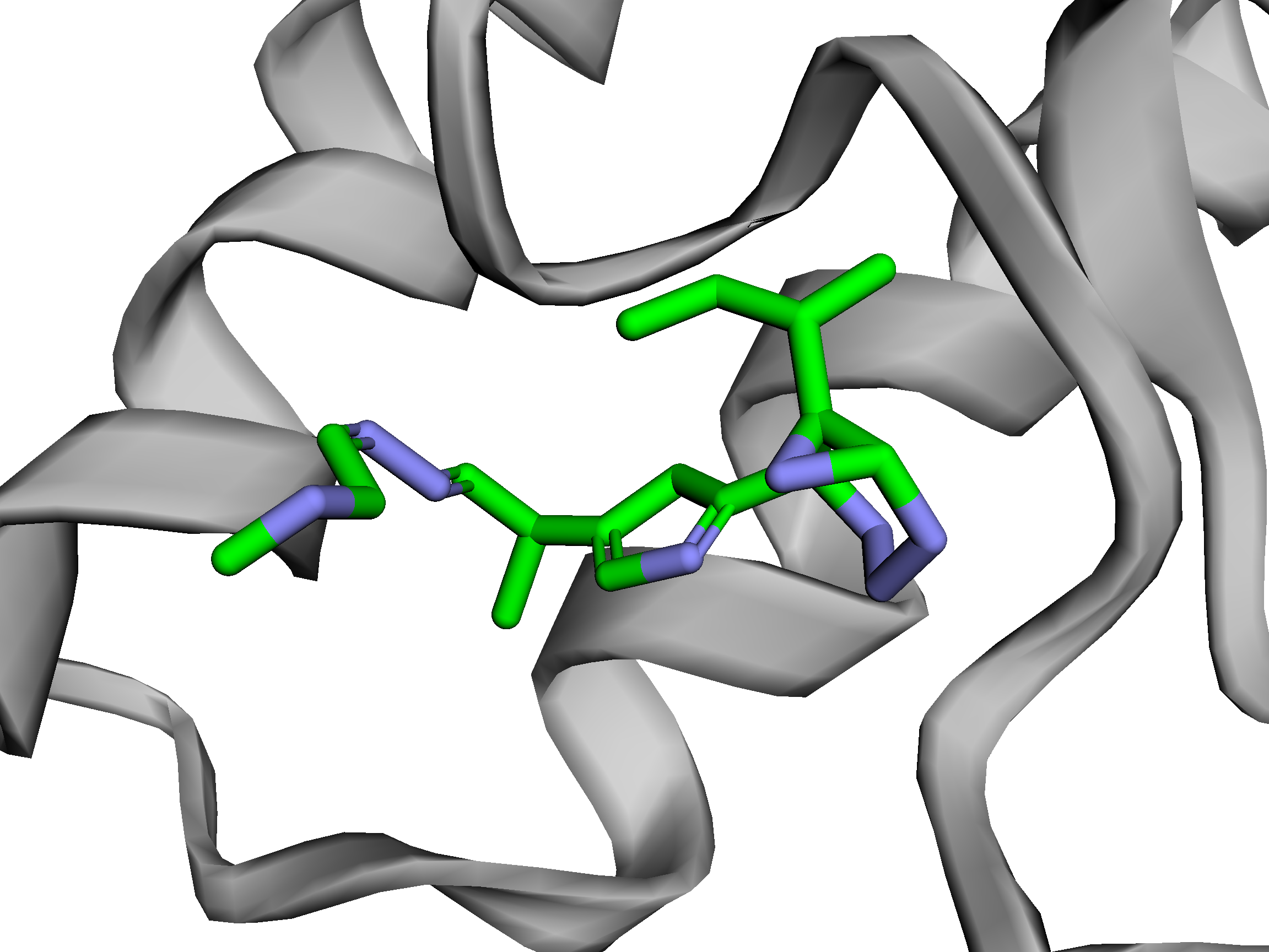}{0.13}
         \\
         \fig{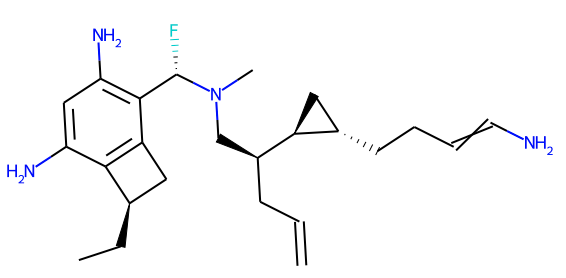}{0.13}
         & \fig{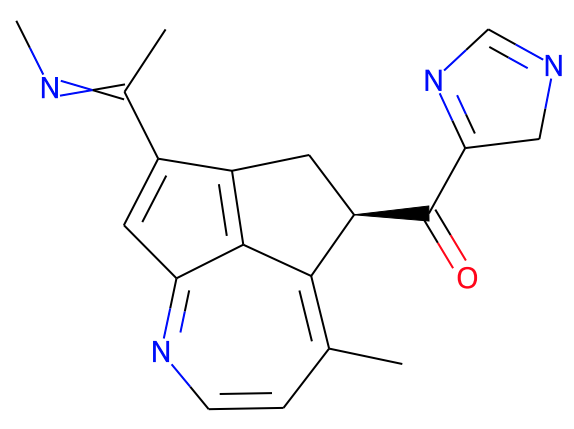}{0.13}
         & \fig{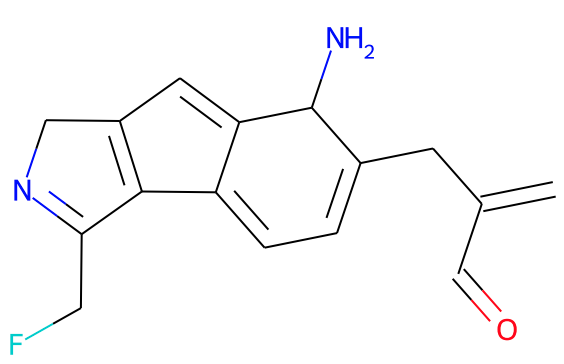}{0.13}
         & \fig{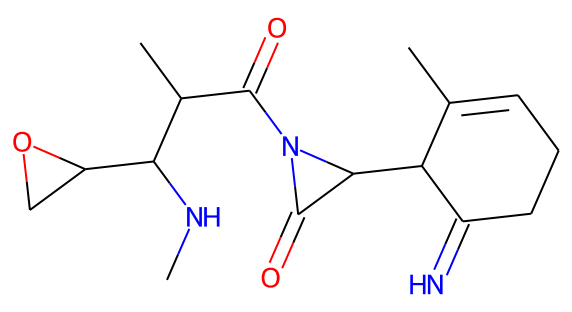}{0.13}
         & \fig{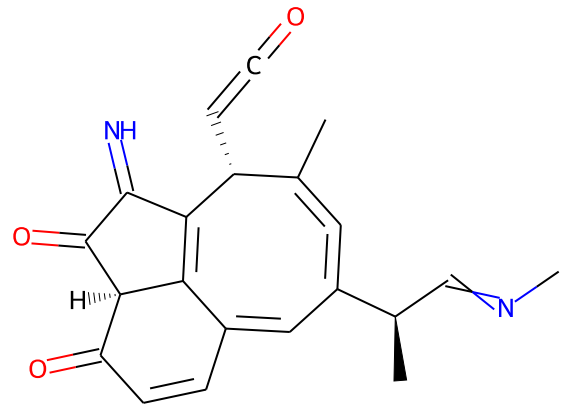}{0.13}
         & \fig{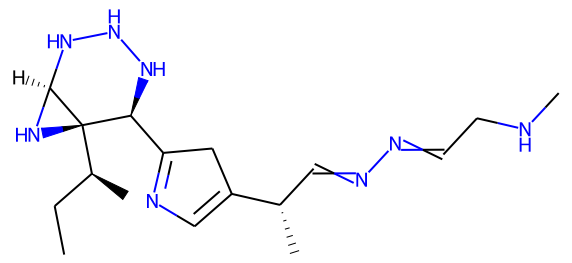}{0.13}
         \\
         \fig{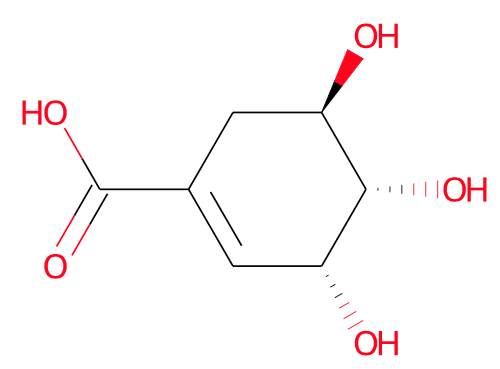}{0.13}
         & \fig{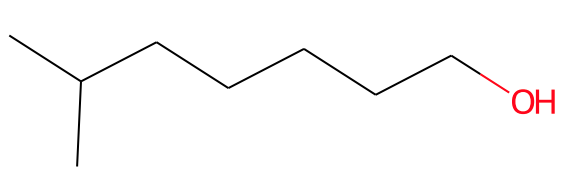}{0.13}
         & \fig{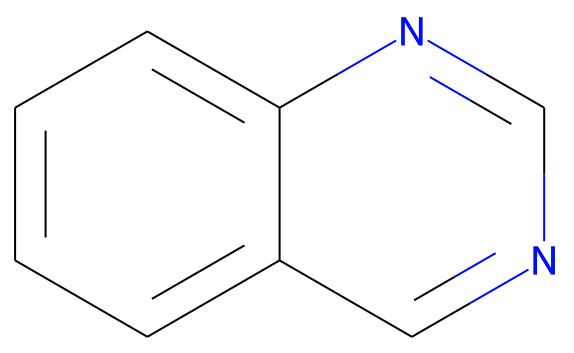}{0.13}
         & \fig{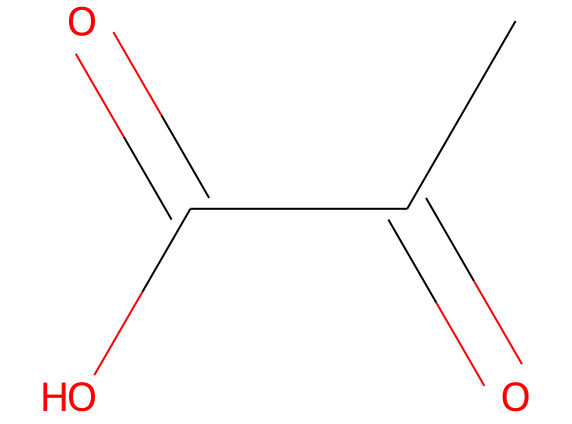}{0.13}
         & \fig{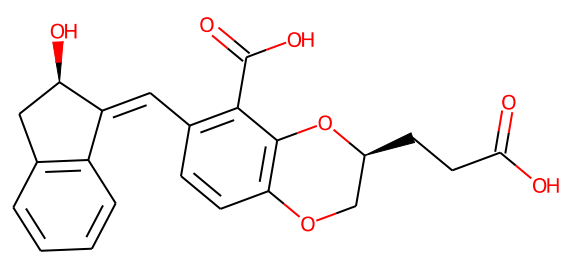}{0.13}
         & \fig{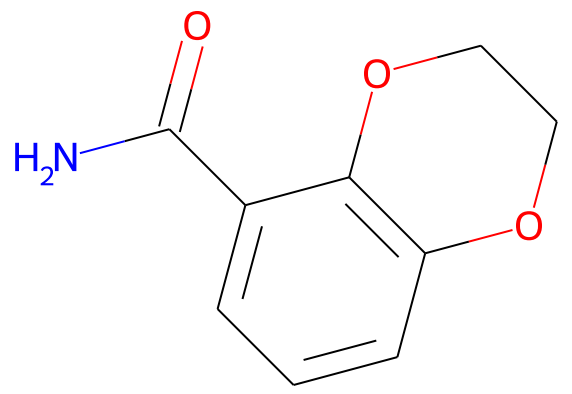}{0.13}
    \end{tabular}
    \caption{Comparison between generated 3D molecules for target binding-site and reference molecules. Receptor IDs, left to right: 1zyu, 2qu9, 2hw1, 3lcg, 3zt3, 5lvq.  Top: generated ligand (colour) + receptor. Middle: generated ligand chemical structure.  Bottom: reference ligand chemical structure.}
    \label{fig:qualitative}
\end{figure*}

%% file: text/conclusions.tex
We have presented a method for learning a conditional distribution of ligands given a receptor.  The method, which is based on a continuous normalizing flow, has provable invariance properties based on semi-equivariance conditions on the flow.  Empirically, our method improves upon competing methods by a considerable margin in $\Delta$Binding, promising the potential to generate previously undiscovered molecules with high binding affinity.

%% file: text/appendix.tex
\subsection{Proof of Theorem}\label{appnx:proof}
In this research we establish semi-equivariance conditions on the continuous graph normalizing flow which guarantee the invariant to rigid body transformations conditions on the conditional distribution.  The theory is described in Section \ref{sec:vertex}, and the proof herein shows formally that the semi-equivariance conditions yield the desired invariant distribution.

\begin{lemma}
\label{lemma:first_transformation}
Let the mean position of the receptor be given by $\bh{x}_{av} = \frac{1}{\hat{N}} \sum_{i=1}^{\hat{N}} \bh{x}_i$, and define the following quantities
\begin{equation*}
    \alpha = \frac{N}{N + \hat{N}} \quad \quad
    \Omega_{\hat{\mathcal{G}}} = 
    \begin{bmatrix}
        \mathbf{I}_{3N} - \frac{\alpha}{N} \mathbf{1}_{N \times N} \otimes \mathbf{I}_3 & \mathbf{0} \\
        \mathbf{0} & \mathbf{I}_{d_h N}
    \end{bmatrix}
    \quad \quad
    \omega_{\hat{\mathcal{G}}} = 
    \begin{bmatrix}
        -(1-\alpha) \mathbf{1}_{N \times 1} \otimes \bh{x}_{av} \\
        \mathbf{0}
    \end{bmatrix}
\end{equation*}
where $\otimes$ indicates the Kronecker product.  Given the following mapping:
\begin{equation}
    \mathbf{v} = \Omega_{\hat{\mathcal{G}}}^{-1} \left( \mathbf{u} - \omega_{\hat{\mathcal{G}}} \right)
\end{equation}
Let the inverse mapping be denoted by $\Gamma_{\hat{\mathcal{G}}}^1$, i.e. $\mathbf{u} = \Gamma_{\hat{\mathcal{G}}}^1(\mathbf{v})$.  For any rigid transformation $T \in E(3)$, which consists of both a rotation and a translation, denote the transformation consisting only of the rotation of $T$ as $T_{rot} \in O(3)$.  Then
\begin{equation*}
    \Gamma_{T\hat{\mathcal{G}}}^1(T\mathbf{v}) = T_{rot} \Gamma_{\hat{\mathcal{G}}}^1(\mathbf{v}).
\end{equation*}
Furthermore, for any permutations $\pi \in \mathbb{S}_N$ and $\hat{\pi} \in \mathbb{S}_{\hat{N}}$, then 
\begin{equation*}
    \Gamma_{\hat{\pi}\hat{\mathcal{G}}}^1(\pi\mathbf{v}) = \pi \Gamma_{\hat{\mathcal{G}}}^1(\mathbf{v}).
\end{equation*}
\end{lemma}

\textbf{\textit{Proof:}} The mapping $\mathbf{u} = \Gamma_{\hat{\mathcal{G}}}^1(\mathbf{v})$ is given by
\begin{equation}
    \mathbf{u} = \Omega_{\hat{\mathcal{G}}} \mathbf{v} + \omega_{\hat{\mathcal{G}}}
\end{equation}
Let us denote the parts of $\mathbf{u}$ corresponding to the coordinates and the features as $\mathbf{x}^\mathbf{u}$ and $\mathbf{h}^\mathbf{u}$, respectively; and use similar notation for $\mathbf{v}$.  Then we have that
\begin{equation}
    \mathbf{h}^\mathbf{u} = \mathbf{h}^\mathbf{v}
    \label{eq:h_trans}
\end{equation}
and
\begin{equation}
    \mathbf{x}^\mathbf{u} = \left( \mathbf{I}_{3N} - \frac{\alpha}{N} \mathbf{1}_{N \times N} \otimes \mathbf{I}_3 \right) \mathbf{x}^\mathbf{v} -(1-\alpha) \mathbf{1}_{N \times 1} \otimes \bh{x}_{av}
\end{equation}
Breaking down this last equation by atom gives
\begin{align}
    \mathbf{x}_i^\mathbf{u}
    & = \mathbf{x}_i^\mathbf{v} - \frac{\alpha}{N} \sum_{j=1}^N \mathbf{x}_j^\mathbf{v} -(1-\alpha) \bh{x}_{av} \notag \\
    & = \mathbf{x}_i^\mathbf{v} - \left( \alpha \mathbf{x}_{av}^\mathbf{v} + (1-\alpha) \bh{x}_{av} \right) \notag \\
    & = \mathbf{x}_i^\mathbf{v} - \bar{\mathbf{x}}^\mathbf{v}
    \label{eq:per_coord_trans}
\end{align}
where $\mathbf{x}_{av}^\mathbf{v}$ is the average coordinate position of $\mathbf{x}^\mathbf{v}$, and $\bar{\mathbf{x}}^\mathbf{v}$ indicates the average of all atoms in the entire complex, i.e. taking both the ligand and the receptor together.

Now, let us examine what happens when we apply the rigid transformation $T$ to both $\mathbf{v}$ and the receptor graph $\hat{\mathcal{G}}$; that is, let us examine
\begin{equation}
    \tilde{\mathbf{u}} = \Gamma_{T\hat{\mathcal{G}}}^1(T\mathbf{v})
\end{equation}
In the case of the features $\mathbf{h}$, they are invariant by design; thus 
\begin{align}
    \mathbf{h}^{\tilde{\mathbf{u}}} 
    & = \mathbf{h}^\mathbf{Tv} \notag \\
    & = \mathbf{h}^\mathbf{v} \notag \\
    & = \mathbf{h}^\mathbf{u}
    \label{eq:u_trans_h}
\end{align}
where the last line follows from Equation (\ref{eq:h_trans}).
In the case of the coordinates, the transformation is as follows:
\begin{equation}
    \mathbf{x}_i^{T\mathbf{v}} = R\mathbf{x}_i^\mathbf{v} + t
\end{equation}
where $R \in O(3)$ is the rotation matrix, and $t \in \mathbb{R}^3$ the translation vector, corresponding to rigid motion $T$.  As we apply $T$ to the receptor graph $\hat{\mathcal{G}}$, this has the effect of applying this transformation to each of the receptor atoms, and hence to their mean and the mean of the entire complex:
\begin{equation}
    \bh{x}_{av}^{T\hat{\mathcal{G}}} = R \bh{x}_{av}^{\hat{\mathcal{G}}} + t \quad \Rightarrow \quad \bar{\mathbf{x}}^\mathbf{Tv} = R \bar{\mathbf{x}}^\mathbf{v} + t
\end{equation}
Thus, following Equation (\ref{eq:per_coord_trans}), and substituting $T\mathbf{v}$ and $T\hat{\mathcal{G}}$ in place of $\mathbf{v}$ and $\hat{\mathcal{G}}$, we get
\begin{align}
    \mathbf{x}_i^{\tilde{\mathbf{u}}}
    & = \mathbf{x}_i^{T\mathbf{v}} - \bar{\mathbf{x}}^{T\mathbf{v}} \notag \\
    & = R\mathbf{x}_i^\mathbf{v} + t - (R\bar{\mathbf{x}}^\mathbf{v} + t ) \notag \\
    & = R \left( \mathbf{x}_i^\mathbf{v} - \bar{\mathbf{x}}^\mathbf{v} \right) \notag \\
    & = R \mathbf{x}_i^\mathbf{u}
    \label{eq:u_trans_x}
\end{align}
Combining Equations (\ref{eq:u_trans_h}) and (\ref{eq:u_trans_x}), we have that
\begin{equation}
    \tilde{\mathbf{u}} = T_{rot} \mathbf{u}
\end{equation}
Since $\mathbf{u} = \Gamma_{\hat{\mathcal{G}}}^1(\mathbf{v})$ and $\tilde{\mathbf{u}} = \Gamma_{T\hat{\mathcal{G}}}^1(T\mathbf{v})$, we have shown that $\Gamma_{T\hat{\mathcal{G}}}^1(T\mathbf{v}) = T_{rot} \Gamma_{\hat{\mathcal{G}}}^1(\mathbf{v})$, as desired. 

In the case of the permutations, let us now set
\begin{equation}
    \tilde{\mathbf{u}} = \Gamma_{\hat{\pi}\hat{\mathcal{G}}}^1(\pi\mathbf{v})
\end{equation}
It is easy to see that $\hat{\pi}$ has no effect; the only place the receptor enters is through the quantities $\hat{N}$ and $\bh{x}_{av}$, both of which are permutation-invariant.  For the features, we now have
\begin{align}
    \mathbf{h}^{\tilde{\mathbf{u}}} 
    & = \mathbf{h}^\mathbf{\pi v} \notag \\
    & = \pi \mathbf{h}^\mathbf{v} \notag \\
    & = \pi \mathbf{h}^\mathbf{u}
\end{align}
That is, the features are simply reordered according to $\pi$.  With regard to the coordinates, we have that
\begin{align}
    \mathbf{x}_i^{\tilde{\mathbf{u}}}
    & = \mathbf{x}_i^{\pi\mathbf{v}} - \bar{\mathbf{x}}^{\pi\mathbf{v}} \notag \\
    & = \mathbf{x}_{\pi(i)}^\mathbf{v} - \bar{\mathbf{x}}^\mathbf{v} \notag \\
    & = \mathbf{x}_{\pi(i)}^\mathbf{u}
\end{align}
The coordinates are also therefore simply reordered according to $\pi$.  Summarizing, we have that
\begin{equation}
    \tilde{\mathbf{u}} = \pi \mathbf{u}
\end{equation}
This is exactly equal to 
\begin{equation*}
    \Gamma_{\hat{\pi}\hat{\mathcal{G}}}^1(\pi\mathbf{v}) = \pi \Gamma_{\hat{\mathcal{G}}}^1(\mathbf{v})
\end{equation*}
which concludes the proof. \qedsymbol

\begin{lemma}
\label{lemma:second_transformation}
Let $\mathbf{u}(1)$ be the output of a Receptor-Conditioned Ligand Flow specified by the Complex-to-Ligand Mapping $\gamma$ which is rotation semi-equivariant and permutation semi-equivariant.  This Receptor-Conditioned Ligand Flow maps the initial condition $\mathbf{z}$ to $\mathbf{u}(1)$; let the inverse mapping be denoted by $\Gamma_{\hat{\mathcal{G}}}^2$, i.e. $\mathbf{z} = \Gamma_{\hat{\mathcal{G}}}^2(\mathbf{u}(1))$.  Then
\begin{equation*}
    \Gamma_{T\hat{\mathcal{G}}}^2(T_{rot} \mathbf{u}) = T_{rot} \Gamma_{\hat{\mathcal{G}}}^2(\mathbf{u})
\end{equation*}
Furthermore, for any permutations $\pi \in \mathbb{S}_N$ and $\hat{\pi} \in \mathbb{S}_{\hat{N}}$, then
\begin{equation*}
    \Gamma_{\hat{\pi}\hat{\mathcal{G}}}^2(\pi \mathbf{u}) = \pi \Gamma_{\hat{\mathcal{G}}}^2(\mathbf{u})
\end{equation*}
\end{lemma}

\textbf{\textit{Proof:}} Our first goal is to show that $\Gamma_{T\hat{\mathcal{G}}}^2(T_{rot} \mathbf{u}) = T_{rot} \Gamma_{\hat{\mathcal{G}}}^2(\mathbf{u})$.  Let us define $F_{\hat{\mathcal{G}}}$ to be the inverse of $\Gamma_{\hat{\mathcal{G}}}^2$, and let $\mathbf{u} = F_{\hat{\mathcal{G}}}(\mathbf{z})$.  Then
\begin{align}
    \Gamma_{T\hat{\mathcal{G}}}^2(T_{rot} \mathbf{u}) = T_{rot} \Gamma_{\hat{\mathcal{G}}}^2(\mathbf{u})
    & \quad \Leftrightarrow \quad F_{T\hat{\mathcal{G}}}^{-1}(T_{rot} F_{\hat{\mathcal{G}}}(\mathbf{z})) = T_{rot} F_{\hat{\mathcal{G}}}^{-1}(F_{\hat{\mathcal{G}}}(\mathbf{z})) \notag \\
    & \quad \Leftrightarrow \quad F_{T\hat{\mathcal{G}}}^{-1}(T_{rot} F_{\hat{\mathcal{G}}}(\mathbf{z})) = T_{rot} \mathbf{z} \notag \\
    & \quad \Leftrightarrow \quad F_{T\hat{\mathcal{G}}}( T_{rot} \mathbf{z} ) = T_{rot} F_{\hat{\mathcal{G}}}(\mathbf{z})
\end{align}
Thus, it is sufficient to shows that $F_{T\hat{\mathcal{G}}}( T_{rot} \mathbf{z} ) = T_{rot} F_{\hat{\mathcal{G}}}(\mathbf{z})$.  For convenience, we shall set
\begin{equation}
    \mathbf{u}(1) = F_{\hat{\mathcal{G}}}(\mathbf{z}) \quad \text{and} \quad \tilde{\mathbf{u}}(1) = F_{T\hat{\mathcal{G}}}( T_{rot} \mathbf{z} )
    \label{eq:u_cond1}
\end{equation}
In this case, $\mathbf{u}(1)$ is defined by the ODE
\begin{equation}
    \frac{d\mathbf{u}}{dt} = \texttt{vec}\left( \gamma(\mathcal{G}_{\mathbf{u}} , \hat{\mathcal{G}}) \right) \quad \text{with } \mathbf{u}(0) = \mathbf{z}
    \label{eq:u_flow1}
\end{equation}
whereas $\tilde{\mathbf{u}}(1)$ is defined by the ODE
\begin{equation}
    \frac{d\tilde{\mathbf{u}}}{dt} = \texttt{vec}\left( \gamma(\mathcal{G}_{\tilde{\mathbf{u}}} , T\hat{\mathcal{G}}) \right) \quad \text{with } \tilde{\mathbf{u}}(0) = T_{rot} \mathbf{z}
    \label{eq:utilde_flow1}
\end{equation}
Now, let us define $\Breve{\mathbf{u}}(t) = T_{rot}^{-1} \tilde{\mathbf{u}}(t)$, so that $\tilde{\mathbf{u}}(t) = T_{rot} \Breve{\mathbf{u}}(t)$.  In this case, we have that:
\begin{enumerate}
    \item $\Breve{\mathbf{u}}(0) = T_{rot}^{-1} \tilde{\mathbf{u}}(0) = T_{rot}^{-1} T_{rot} \mathbf{z} = \mathbf{z}$.
    \item $\frac{d\tilde{\mathbf{u}}}{dt} = T_{rot} \frac{d\Breve{\mathbf{u}}}{dt}$.
    \item $\texttt{vec}\left( \gamma(\mathcal{G}_{\tilde{\mathbf{u}}} , T\hat{\mathcal{G}}) \right) = \texttt{vec}\left( \gamma(\mathcal{G}_{T_{rot}\Breve{\mathbf{u}}} , T\hat{\mathcal{G}}) \right) = T_{rot} \texttt{vec}\left( \gamma(\mathcal{G}_{\Breve{\mathbf{u}}} , \hat{\mathcal{G}}) \right)$, where the last equality is from the definition of rotation semi-equivariance of $\gamma$.
\end{enumerate}
Plugging the above three results into the flow for $\tilde{\mathbf{u}}$ in Equation (\ref{eq:utilde_flow1}) yields
\begin{align}
    & T_{rot} \frac{d\Breve{\mathbf{u}}}{dt}  = T_{rot} \texttt{vec}\left( \gamma(\mathcal{G}_{\Breve{\mathbf{u}}} , \hat{\mathcal{G}}) \right) \quad \text{with} \quad \Breve{\mathbf{u}}(0) = \mathbf{z} \notag \\
    & \Rightarrow \quad \frac{d\Breve{\mathbf{u}}}{dt}  = \texttt{vec}\left( \gamma(\mathcal{G}_{\Breve{\mathbf{u}}} , \hat{\mathcal{G}}) \right) \quad \text{with} \quad \Breve{\mathbf{u}}(0) = \mathbf{z}
\end{align}
But this is precisely identical to the flow described in Equation (\ref{eq:u_flow1}); thus, we have that
\begin{equation}
    \Breve{\mathbf{u}}(t) = \mathbf{u}(t) \quad \text{for all } t
\end{equation}
But $\tilde{\mathbf{u}}(t) = T_{rot} \Breve{\mathbf{u}}(t)$ so that $\tilde{\mathbf{u}}(t) = T_{rot} \mathbf{u}(t)$, and in particular $\tilde{\mathbf{u}}(1) = T_{rot} \mathbf{u}(1)$.  Comparing with Equation (\ref{eq:u_cond1}) completes rigid motion part of the proof.

Let us now turn to permutations; the proof is similar, but we repeat it in full for completeness.  Our goal is to show that $\Gamma_{\hat{\pi}\hat{\mathcal{G}}}^2(\pi \mathbf{u}) = \pi \Gamma_{\hat{\mathcal{G}}}^2(\mathbf{u})$.  Let us define $F_{\hat{\mathcal{G}}}$ to be the inverse of $\Gamma_{\hat{\mathcal{G}}}^2$, and let $\mathbf{u} = F_{\hat{\mathcal{G}}}(\mathbf{z})$.  Then
\begin{align}
    \Gamma_{\hat{\pi}\hat{\mathcal{G}}}^2(\pi \mathbf{u}) = \pi \Gamma_{\hat{\mathcal{G}}}^2(\mathbf{u})
    & \quad \Leftrightarrow \quad F_{\hat{\pi}\hat{\mathcal{G}}}^{-1}(\pi F_{\hat{\mathcal{G}}}(\mathbf{z})) = \pi F_{\hat{\mathcal{G}}}^{-1}(F_{\hat{\mathcal{G}}}(\mathbf{z})) \notag \\
    & \quad \Leftrightarrow \quad F_{\hat{\pi}\hat{\mathcal{G}}}^{-1}(\pi F_{\hat{\mathcal{G}}}(\mathbf{z})) = \pi \mathbf{z} \notag \\
    & \quad \Leftrightarrow \quad F_{\hat{\pi}\hat{\mathcal{G}}}( \pi \mathbf{z} ) = \pi F_{\hat{\mathcal{G}}}(\mathbf{z})
\end{align}
Thus, it is sufficient to shows that $F_{\hat{\pi}\hat{\mathcal{G}}}( \pi \mathbf{z} ) = \pi F_{\hat{\mathcal{G}}}(\mathbf{z})$.  For convenience, we shall set
\begin{equation}
    \mathbf{u}(1) = F_{\hat{\mathcal{G}}}(\mathbf{z}) \quad \text{and} \quad \tilde{\mathbf{u}}(1) = F_{\hat{\pi}\hat{\mathcal{G}}}( \pi \mathbf{z} )
    \label{eq:u_cond}
\end{equation}
In this case, $\mathbf{u}(1)$ is defined by the ODE
\begin{equation}
    \frac{d\mathbf{u}}{dt} = \texttt{vec}\left( \gamma(\mathcal{G}_{\mathbf{u}} , \hat{\mathcal{G}}) \right) \quad \text{with } \mathbf{u}(0) = \mathbf{z}
    \label{eq:u_flow}
\end{equation}
whereas $\tilde{\mathbf{u}}(1)$ is defined by the ODE
\begin{equation}
    \frac{d\tilde{\mathbf{u}}}{dt} = \texttt{vec}\left( \gamma(\mathcal{G}_{\tilde{\mathbf{u}}} , \hat{\pi}\hat{\mathcal{G}}) \right) \quad \text{with } \tilde{\mathbf{u}}(0) = \pi \mathbf{z}
    \label{eq:utilde_flow}
\end{equation}
Now, let us define $\Breve{\mathbf{u}}(t) = \pi^{-1} \tilde{\mathbf{u}}(t)$, so that $\tilde{\mathbf{u}}(t) = \pi \Breve{\mathbf{u}}(t)$.  In this case, we have that:
\begin{enumerate}
    \item $\Breve{\mathbf{u}}(0) = \pi^{-1} \tilde{\mathbf{u}}(0) = \pi^{-1} \pi \mathbf{z} = \mathbf{z}$.
    \item $\frac{d\tilde{\mathbf{u}}}{dt} = \pi \frac{d\Breve{\mathbf{u}}}{dt}$.
    \item $\texttt{vec}\left( \gamma(\mathcal{G}_{\tilde{\mathbf{u}}} , \hat{\pi}\hat{\mathcal{G}}) \right) = \texttt{vec}\left( \gamma(\mathcal{G}_{\pi\Breve{\mathbf{u}}} , \hat{\pi}\hat{\mathcal{G}}) \right) = \pi \texttt{vec}\left( \gamma(\mathcal{G}_{\Breve{\mathbf{u}}} , \hat{\mathcal{G}}) \right)$, where the last equality is from the definition of permutation semi-equivariance of $\gamma$.
\end{enumerate}
Plugging the above three results into the flow for $\tilde{\mathbf{u}}$ in Equation (\ref{eq:utilde_flow}) yields
\begin{align}
    & \pi \frac{d\Breve{\mathbf{u}}}{dt}  = \pi \texttt{vec}\left( \gamma(\mathcal{G}_{\Breve{\mathbf{u}}} , \hat{\mathcal{G}}) \right) \quad \text{with} \quad \Breve{\mathbf{u}}(0) = \mathbf{z} \notag \\
    & \Rightarrow \quad \frac{d\Breve{\mathbf{u}}}{dt}  = \texttt{vec}\left( \gamma(\mathcal{G}_{\Breve{\mathbf{u}}} , \hat{\mathcal{G}}) \right) \quad \text{with} \quad \Breve{\mathbf{u}}(0) = \mathbf{z}
\end{align}
But this is precisely identical to the flow described in Equation (\ref{eq:u_flow}); thus, we have that
\begin{equation}
    \Breve{\mathbf{u}}(t) = \mathbf{u}(t) \quad \text{for all } t
\end{equation}
But $\tilde{\mathbf{u}}(t) = \pi \Breve{\mathbf{u}}(t)$ so that $\tilde{\mathbf{u}}(t) = \pi \mathbf{u}(t)$, and in particular $\tilde{\mathbf{u}}(1) = \pi \mathbf{u}(1)$.  Comparing with Equation (\ref{eq:u_cond}) completes the proof. \qedsymbol

\begin{theorem}
Let $\mathbf{u}(1)$ be the output of a Receptor-Conditioned Ligand Flow specified by the Complex-to-Ligand Mapping $\gamma$.  Let the mean position of the receptor be given by $\bh{x}_{av} = \frac{1}{\hat{N}} \sum_{i=1}^{\hat{N}} \bh{x}_i$, and define the following quantities
\begin{equation}
    \alpha = \frac{N}{N + \hat{N}} \quad \quad
    \Omega_{\hat{\mathcal{G}}} = 
    \begin{bmatrix}
        \mathbf{I}_{3N} - \frac{\alpha}{N} \mathbf{1}_{N \times N} \otimes \mathbf{I}_3 & \mathbf{0} \\
        \mathbf{0} & \mathbf{I}_{d_h N}
    \end{bmatrix}
    \quad \quad
    \omega_{\hat{\mathcal{G}}} = 
    \begin{bmatrix}
        -(1-\alpha) \mathbf{1}_{N \times 1} \otimes \bh{x}_{av} \\
        \mathbf{0}
    \end{bmatrix}
\end{equation}
where $\otimes$ indicates the Kronecker product.  Finally, let 
\begin{equation}
    \mathbf{v} = \Omega_{\hat{\mathcal{G}}}^{-1} \left( \mathbf{u}(1) - \omega_{\hat{\mathcal{G}}} \right)
    \label{eq:final_transformation}
\end{equation}

Suppose that $\gamma$ is both rotation semi-equivariant and permutation semi-equivariant.  Then the resulting distribution on $\mathbf{v}$, that is $p_{vec}(\mathbf{v} | \hat{\mathcal{G}})$, yields a vertex distribution $p(V | N, \hat{\mathcal{G}}) = p_{vec}(\texttt{vec}(V) | \hat{\mathcal{G}})$ that satisfies the invariance conditions in Equation (\ref{eq:vertex_invariance}).
\end{theorem}

\textbf{\textit{Proof:}} The Receptor-Conditioned Ligand Flow maps from the Gaussian random variable $\mathbf{z}$ to the variable $\mathbf{u}(1)$.  As this flow is a normalizing flow, it is invertible, so let us denote the inverse mapping by $\Gamma^2$:
\begin{equation}
    \mathbf{z} = \Gamma_{\hat{\mathcal{G}}}^2(\mathbf{u}(1))
\end{equation}
Note that the dependence on the receptor graph $\hat{\mathcal{G}}$ is denoted using a subscript, as the invertibility does not apply to the receptor, but only to the ligand.  Equation (\ref{eq:final_transformation}) maps from the variable $\mathbf{u}(1)$ to the variable $\mathbf{v}$; let us denote its inverse mapping by $\Gamma^1$:
\begin{equation}
    \mathbf{u}(1) = \Gamma_{\hat{\mathcal{G}}}^1(\mathbf{v})
\end{equation}
In this case, we have that
\begin{equation}
    \mathbf{z} = \Gamma_{\hat{\mathcal{G}}}^2(\Gamma_{\hat{\mathcal{G}}}^1(\mathbf{v})) \equiv \Gamma_{\hat{\mathcal{G}}}(\mathbf{v})
    \label{eq:v_to_z}
\end{equation}

Now, our goal is to show that the following condition holds:
\begin{equation}
    p(TV | N, T\hat{\mathcal{G}}) = p(V | N, \hat{\mathcal{G}}) \hspace{0.5cm} \text{for } T \in E(3)
\end{equation}
Using the $p_{vec}$ notation, this translates to
\begin{equation}
    p_{vec}(T\mathbf{v} | T\hat{\mathcal{G}}) = p_{vec}(\mathbf{v} | \hat{\mathcal{G}})
    \label{eq:invariance_with_vec}
\end{equation}
Now, from Equation (\ref{eq:v_to_z}), the fact that $\Gamma$ is invertible, and the change of variables formula, we have that
\begin{equation}
    p_{vec}(\mathbf{v} | \hat{\mathcal{G}}) = p_{\mathbf{z}}(\Gamma_{\hat{\mathcal{G}}}(\mathbf{v})) |\det J_{\Gamma_{\hat{\mathcal{G}}}}(\mathbf{v}) |
\end{equation}
where $p_{\mathbf{z}}(\cdot)$ is the Gaussian distribution from $\mathbf{z}$ is sampled; and $J_{\Gamma_{\hat{\mathcal{G}}}}(\cdot)$ is the Jacobian of $\Gamma_{\hat{\mathcal{G}}}(\cdot)$.  Since $\Gamma_{\hat{\mathcal{G}}} = \Gamma_{\hat{\mathcal{G}}}^2 \circ \Gamma_{\hat{\mathcal{G}}}^1$, this can be expanded as
\begin{equation}
    p_{vec}(\mathbf{v} | \hat{\mathcal{G}}) = p_{\mathbf{z}}( \Gamma_{\hat{\mathcal{G}}}^2(\Gamma_{\hat{\mathcal{G}}}^1(\mathbf{v})) ) |\det J_{\Gamma_{\hat{\mathcal{G}}}^2}(\Gamma_{\hat{\mathcal{G}}}^1(\mathbf{v})) | |\det J_{\Gamma_{\hat{\mathcal{G}}}^1}(\mathbf{v}) |
\end{equation}
using the chain rule, and the fact that determinant of a product is the product of determinants.  Plugging this into Equation (\ref{eq:invariance_with_vec}), we must show that
\begin{align}
    & p_{\mathbf{z}}( \Gamma_{T\hat{\mathcal{G}}}^2(\Gamma_{T\hat{\mathcal{G}}}^1(T\mathbf{v})) ) |\det J_{\Gamma_{T\hat{\mathcal{G}}}^2}(\Gamma_{T\hat{\mathcal{G}}}^1(T\mathbf{v})) | |\det J_{\Gamma_{T\hat{\mathcal{G}}}^1}(T\mathbf{v}) |  \notag \\
    & \hspace{1.0cm} = \quad p_{\mathbf{z}}( \Gamma_{\hat{\mathcal{G}}}^2(\Gamma_{\hat{\mathcal{G}}}^1(\mathbf{v})) ) |\det J_{\Gamma_{\hat{\mathcal{G}}}^2}(\Gamma_{\hat{\mathcal{G}}}^1(\mathbf{v})) | |\det J_{\Gamma_{\hat{\mathcal{G}}}^1}(\mathbf{v}) |
    \hspace{0.5cm} \text{for } T \in E(3)
    \label{eq:to_prove}
\end{align}
A rigid transformation $T \in E(3)$ consists of both a rotation and a translation.  For brevity, denote the transformation consisting only of the rotation of $T$ as $T_{rot} \in O(3)$.  Now, from Lemma \ref{lemma:first_transformation}, we have that
\begin{equation}
    \Gamma_{T\hat{\mathcal{G}}}^1(T\mathbf{v}) = T_{rot} \Gamma_{\hat{\mathcal{G}}}^1(\mathbf{v})
    \label{eq:lemma_repr_a}
\end{equation}
From Lemma \ref{lemma:second_transformation}, we have that
\begin{equation}
    \Gamma_{T\hat{\mathcal{G}}}^2(T_{rot} \mathbf{u}) = T_{rot} \Gamma_{\hat{\mathcal{G}}}^2(\mathbf{u})
    \label{eq:lemma_repr_b}
\end{equation}
Combining Equations (\ref{eq:lemma_repr_a}) and (\ref{eq:lemma_repr_b}) gives that
\begin{align}
    p_{\mathbf{z}}( \Gamma_{T\hat{\mathcal{G}}}^2(\Gamma_{T\hat{\mathcal{G}}}^1(T \mathbf{v})) )
    & = p_{\mathbf{z}}( \Gamma_{T\hat{\mathcal{G}}}^2(T_{rot}\Gamma_{\hat{\mathcal{G}}}^1(\mathbf{v})) ) \notag \\
    & = p_{\mathbf{z}}( T_{rot} \Gamma_{\hat{\mathcal{G}}}^2(\Gamma_{\hat{\mathcal{G}}}^1(\mathbf{v})) ) \notag \\
    & = p_{\mathbf{z}}( \Gamma_{\hat{\mathcal{G}}}^2(\Gamma_{\hat{\mathcal{G}}}^1(\mathbf{v})) )
    \label{eq:distribution_result_1}
\end{align}
where the last line follows from the rotation invariance of the Gaussian distribution.

Note that
\begin{align}
    J_{\Gamma_{T\hat{\mathcal{G}}}^1}(T\mathbf{v})
    & = \frac{\partial}{\partial \mathbf{v}} \left( \Gamma_{T\hat{\mathcal{G}}}^1 (T\mathbf{v}) \right) \notag \\
    & = \frac{\partial}{\partial \mathbf{v}} \left( T_{rot}\Gamma_{\hat{\mathcal{G}}}^1(\mathbf{v}) \right) \notag \\
    & = T_{rot} \frac{\partial}{\partial \mathbf{v}} \left( \Gamma_{\hat{\mathcal{G}}}^1(\mathbf{v}) \right) \notag \\
    & = T_{rot} J_{\Gamma_{\hat{\mathcal{G}}}^1}(\mathbf{v})
    \label{eq:JGamma1}
\end{align}
Now, $T_{rot}$ can be represented by the $d_v^N \times d_v^N$ block diagonal matrix given by
\begin{equation}
    T_{rot} =
    \begin{bmatrix}
        \mathbf{1}_{N \times 1} \otimes R & 0 \\
        0 & \mathbf{I}_{d_h N}
    \end{bmatrix}
\end{equation}
where $R \in O(3)$, the top-left block corresponds to the coordinates $\mathbf{x}$ and the bottom-right block corresponds to the feature $\mathbf{h}$.  Thus,
\begin{align}
    \det( J_{\Gamma_{T\hat{\mathcal{G}}}^1}(T\mathbf{v}) )
    & = \det( T_{rot} J_{\Gamma_{\hat{\mathcal{G}}}^1}(\mathbf{v}) ) \notag \\
    & = \det( T_{rot} ) \det( J_{\Gamma_{\hat{\mathcal{G}}}^1}(\mathbf{v}) ) \notag \\
    & = \det(R)^N \det(\mathbf{I}_{d_h N}) \det( J_{\Gamma_{\hat{\mathcal{G}}}^1}(\mathbf{v}) ) \notag \\
    & = \pm \det( J_{\Gamma_{\hat{\mathcal{G}}}^1}(\mathbf{v}) )
    \label{eq:jacobian_result_1}
\end{align}
where the second line follows from the fact that the determinant of a product is the product of determinants; the third line from the fact that the determinant of a block diagonal matrix is the product of the determinants of the blocks; and the fourth line from the fact that the determinant of a rotation matrix is $\pm 1$.

To simplify $J_{\Gamma_{T\hat{\mathcal{G}}}^2}(\Gamma_{T\hat{\mathcal{G}}}^1(T\mathbf{v}))$, note that
\begin{align}
    \Gamma_{T\hat{\mathcal{G}}}^2(\mathbf{u})
    & = \Gamma_{T\hat{\mathcal{G}}}^2(T_{rot} T_{rot}^{-1} \mathbf{u}) \notag \\
    & = T_{rot} \Gamma_{\hat{\mathcal{G}}}^2( T_{rot}^{-1} \mathbf{u})
\end{align}
where we have used Equation (\ref{eq:lemma_repr_b}).  Thus,
\begin{align}
    J_{\Gamma_{T\hat{\mathcal{G}}}^2}(\mathbf{u})
    & = \frac{\partial}{\partial \mathbf{u}} \left( \Gamma_{T\hat{\mathcal{G}}}^2(\mathbf{u}) \right) \notag \\
    & = \frac{\partial}{\partial \mathbf{u}} \left( T_{rot} \Gamma_{\hat{\mathcal{G}}}^2( T_{rot}^{-1} \mathbf{u}) \right) \notag \\
    & = T_{rot} J_{\Gamma_{T\hat{\mathcal{G}}}^2}( T_{rot}^{-1} \mathbf{u} ) T_{rot}^{-1}
    \label{eq:jac_simp}
\end{align}
We wish to plug in $\mathbf{u} = \Gamma_{T\hat{\mathcal{G}}}^1(T\mathbf{v})$.  Note that from Equation (\ref{eq:lemma_repr_a}), $\Gamma_{T\hat{\mathcal{G}}}^1(T\mathbf{v}) = T_{rot} \Gamma_{\hat{\mathcal{G}}}^1(\mathbf{v})$.  Thus,
\begin{align}
    J_{\Gamma_{T\hat{\mathcal{G}}}^2}(\Gamma_{T\hat{\mathcal{G}}}^1(T\mathbf{v}))
    & = J_{\Gamma_{T\hat{\mathcal{G}}}^2}( T_{rot} \Gamma_{\hat{\mathcal{G}}}^1(\mathbf{v}) ) \notag \\
    & = T_{rot} J_{\Gamma_{T\hat{\mathcal{G}}}^2}( T_{rot}^{-1} T_{rot} \Gamma_{\hat{\mathcal{G}}}^1(\mathbf{v}) ) T_{rot}^{-1} \notag \\
    & = T_{rot} J_{\Gamma_{T\hat{\mathcal{G}}}^2}( \Gamma_{\hat{\mathcal{G}}}^1(\mathbf{v}) ) T_{rot}^{-1}
    \label{eq:JGamma2}
\end{align}
where in the second line we substituted Equation (\ref{eq:jac_simp}).  Taking determinants gives
\begin{align}
    \det \left( J_{\Gamma_{T\hat{\mathcal{G}}}^2}(\Gamma_{T\hat{\mathcal{G}}}^1(T\mathbf{v})) \right)
    & = \det \left( T_{rot} J_{\Gamma_{T\hat{\mathcal{G}}}^2}( \Gamma_{\hat{\mathcal{G}}}^1(\mathbf{v}) ) T_{rot}^{-1} \right) \notag \\
    & = \det \left( T_{rot}^{-1} T_{rot} J_{\Gamma_{T\hat{\mathcal{G}}}^2}( \Gamma_{\hat{\mathcal{G}}}^1(\mathbf{v}) )  \right) \notag \\
    & = \det \left( J_{\Gamma_{T\hat{\mathcal{G}}}^2}( \Gamma_{\hat{\mathcal{G}}}^1(\mathbf{v}) )  \right)
    \label{eq:jacobian_result_2}
\end{align}
where in the second line, we used the fact that permuting the order of a matrix multiplication does not affect the determinant.

Combining Equations (\ref{eq:distribution_result_1}), (\ref{eq:jacobian_result_1}), and (\ref{eq:jacobian_result_2}), we finally arrive at:
\begin{align}
    & p_{\mathbf{z}}( \Gamma_{T\hat{\mathcal{G}}}^2(\Gamma_{T\hat{\mathcal{G}}}^1(T\mathbf{v})) ) |\det J_{\Gamma_{T\hat{\mathcal{G}}}^2}(\Gamma_{T\hat{\mathcal{G}}}^1(T\mathbf{v})) | |\det J_{\Gamma_{T\hat{\mathcal{G}}}^1}(T\mathbf{v}) |  \notag \\
    & \hspace{1.0cm} = \quad p_{\mathbf{z}}( \Gamma_{\hat{\mathcal{G}}}^2(\Gamma_{\hat{\mathcal{G}}}^1(\mathbf{v})) ) |\det J_{\Gamma_{\hat{\mathcal{G}}}^2}(\Gamma_{\hat{\mathcal{G}}}^1(\mathbf{v})) | |\det J_{\Gamma_{\hat{\mathcal{G}}}^1}(\mathbf{v}) |
\end{align}
which is exactly Equation (\ref{eq:to_prove}).  Thus, we have shown that $p(TV | N, T\hat{\mathcal{G}}) = p(V | N, \hat{\mathcal{G}})$, as desired.  

Let us turn now to the permutation case, which is quite similar.  Similar to Equation (\ref{eq:to_prove}), we need to show
\begin{align}
    & p_{\mathbf{z}}( \Gamma_{\hat{\pi}\hat{\mathcal{G}}}^2(\Gamma_{\hat{\pi}\hat{\mathcal{G}}}^1(\pi\mathbf{v})) ) |\det J_{\Gamma_{\hat{\pi}\hat{\mathcal{G}}}^2}(\Gamma_{\hat{\pi}\hat{\mathcal{G}}}^1(\pi\mathbf{v})) | |\det J_{\Gamma_{\hat{\pi}\hat{\mathcal{G}}}^1}(\pi\mathbf{v}) |  \notag \\
    & \hspace{1.0cm} = \quad p_{\mathbf{z}}( \Gamma_{\hat{\mathcal{G}}}^2(\Gamma_{\hat{\mathcal{G}}}^1(\mathbf{v})) ) |\det J_{\Gamma_{\hat{\mathcal{G}}}^2}(\Gamma_{\hat{\mathcal{G}}}^1(\mathbf{v})) | |\det J_{\Gamma_{\hat{\mathcal{G}}}^1}(\mathbf{v}) |
    \hspace{0.5cm} \text{for } \pi \in \mathbb{S}_n \text{ and } \hat{\pi} \in \mathbb{S}_{\hat{N}}
    \label{eq:perm_to_prove}
\end{align}
From Lemmata 1 and 2, we have that
\begin{equation}
    \Gamma_{\hat{\pi}\hat{\mathcal{G}}}^1(\pi\mathbf{v}) = \pi \Gamma_{\hat{\mathcal{G}}}^1(\mathbf{v}) \quad \text{and} \quad \Gamma_{\hat{\pi}\hat{\mathcal{G}}}^2(\pi \mathbf{u}) = \pi \Gamma_{\hat{\mathcal{G}}}^2(\mathbf{u})
\end{equation}
Thus
\begin{align}
    p_{\mathbf{z}}( \Gamma_{\hat{\pi}\hat{\mathcal{G}}}^2(\Gamma_{\hat{\pi}\hat{\mathcal{G}}}^1(\pi\mathbf{v})) )
    & = p_{\mathbf{z}}( \Gamma_{\hat{\pi}\hat{\mathcal{G}}}^2(\pi \Gamma_{\hat{\mathcal{G}}}^1(\mathbf{v})) ) \notag \\
    & = p_{\mathbf{z}}( \pi \Gamma_{\hat{\mathcal{G}}}^2( \Gamma_{\hat{\mathcal{G}}}^1(\mathbf{v})) ) \notag \\
    & = p_{\mathbf{z}}( \Gamma_{\hat{\mathcal{G}}}^2( \Gamma_{\hat{\mathcal{G}}}^1(\mathbf{v})) )
    \label{eq:perm_result_1}
\end{align}
where the last line follows from the permutation-invariance of the Gaussian distribution $ p_{\mathbf{z}}(\cdot)$.

In a manner parallel to the derivation of Equation (\ref{eq:JGamma1}), we can show that
\begin{equation}
    J_{\Gamma_{\hat{\pi}\hat{\mathcal{G}}}^1}(\pi\mathbf{v}) = \boldsymbol{\pi} J_{\Gamma_{\hat{\mathcal{G}}}^1}(\mathbf{v})
\end{equation}
where $\boldsymbol{\pi}$ now indicates the permutation matrix associated with the permutation $\pi$.  Thus, we have that
\begin{equation}
    \det\left( J_{\Gamma_{\hat{\pi}\hat{\mathcal{G}}}^1}(\pi\mathbf{v}) \right) = \det(\boldsymbol{\pi}) \det\left( J_{\Gamma_{\hat{\mathcal{G}}}^1}(\mathbf{v}) \right) = \pm \det\left( J_{\Gamma_{\hat{\mathcal{G}}}^1}(\mathbf{v}) \right)
    \label{eq:perm_result_2}
\end{equation}
where we have used the fact that a permutation matrix has determinant of $\pm 1$.  Similarly, in a manner parallel to the derivation of Equation (\ref{eq:JGamma2}), we can show that 
\begin{equation}
    J_{\Gamma_{\hat{\pi}\hat{\mathcal{G}}}^2}(\Gamma_{\hat{\pi}\hat{\mathcal{G}}}^1(\pi\mathbf{v})) = \boldsymbol{\pi} J_{\Gamma_{\hat{\mathcal{G}}}^2}(\Gamma_{\hat{\mathcal{G}}}^1(\mathbf{v})) \boldsymbol{\pi}^{-1}
\end{equation}
so that
\begin{align}
    \det\left( J_{\Gamma_{\hat{\pi}\hat{\mathcal{G}}}^2}(\Gamma_{\hat{\pi}\hat{\mathcal{G}}}^1(\pi\mathbf{v})) \right)
    & = \det\left( \boldsymbol{\pi} J_{\Gamma_{\hat{\mathcal{G}}}^2}(\Gamma_{\hat{\mathcal{G}}}^1(\mathbf{v})) \boldsymbol{\pi}^{-1} \right) \notag \\
    & = \det\left( \boldsymbol{\pi}^{-1} \boldsymbol{\pi} J_{\Gamma_{\hat{\mathcal{G}}}^2}(\Gamma_{\hat{\mathcal{G}}}^1(\mathbf{v})) \right) \notag \\
    & = \det\left( J_{\Gamma_{\hat{\mathcal{G}}}^2}(\Gamma_{\hat{\mathcal{G}}}^1(\mathbf{v})) \right)
    \label{eq:perm_result_3}
\end{align}
Combining Equations (\ref{eq:perm_result_1}), (\ref{eq:perm_result_2}), and (\ref{eq:perm_result_3}) yields Equation (\ref{eq:perm_to_prove}); completing the proof.  \qedsymbol

\subsection{Data}\label{appnx:data}
For empirical testing of the method, we use the CrossDocked2020 dataset \citep{francoeur2020three}, with a data-split into a training set (see \href{https://bits.csb.pitt.edu/files/it2_tt_0_lowrmsd_mols_train0_fixed.types}{\texttt{here}}) and validation set (see \href{https://bits.csb.pitt.edu/files/it2_tt_0_lowrmsd_mols_test0_fixed.types}{\texttt{here}}); with additional filtering, as described in Section \ref{sec:experiments}. The evaluation set is given \href{https://github.com/mattragoza/LiGAN/blob/master/data/crossdock2020/selected_test_targets.types}{\texttt{here}}, and we use the same approach to evaluate the generative model as done in previous works \citep{ragoza2022generating, liu2022generating}. The reference evaluation set consists of 10 target receptors with each having multiple associated ligands, leading to 90 receptor-ligand pairs. After the previously described filtering procedure (Sec. \ref{sec:experiments}), 5 receptors remain corresponding to 27 receptor-ligand pairs. We complete the evaluation set, to yield 10 receptors corresponding to 90 pairs, by selecting data points from the validation set. These data points are chosen so that the ligands, receptors and pockets do not appear in the training set, and are not repeated in the evaluation set. In order to add bonds to the resulting molecules, we rely on the LiGAN implementation, see \href{https://github.com/mattragoza/LiGAN/blob/master/ligan/bond_adding.py}{\texttt{here}}


